%% file: acl_latex.tex
\documentclass[11pt]{article}

\usepackage[final]{acl}
\usepackage{times}
\usepackage{latexsym}
\usepackage[T1]{fontenc}
\usepackage[utf8]{inputenc}
\usepackage{microtype}
\usepackage{inconsolata}
\usepackage{graphicx}
\usepackage{booktabs}
\usepackage{amsmath}
\usepackage{amssymb}
\usepackage{bbm}
\usepackage{array}
\usepackage{tabularx}
\usepackage{enumitem}
\usepackage{xspace}
\usepackage{fontawesome5}

\usepackage[most]{tcolorbox}

\newtcblisting{promptboxcol}[1]{
  enhanced,
  listing only,
  listing engine=listings,
  listing options={
    basicstyle=\scriptsize\ttfamily,
    breaklines=true,
    columns=fullflexible,
    keepspaces=true,
    showstringspaces=false
  },
  colback=black!3,
  colframe=black!70,
  colbacktitle=black!75,
  coltitle=white,
  fonttitle=\bfseries,
  title={#1},
  boxrule=0.5pt,
  arc=2pt,
  left=4pt,
  right=4pt,
  top=4pt,
  bottom=4pt,
  before skip=4pt,
  after skip=6pt
}

\newtcolorbox{questionbox}{
  colback=gray!10,
  colframe=gray!50,
  boxrule=1.5pt,
  arc=2pt,
  left=4pt,
  right=4pt,
  top=3pt,
  bottom=3pt,
  boxsep=2pt,
  before skip=4pt,
  after skip=4pt,
}

\definecolor{todoblue}{RGB}{0,102,255}

\newcommand{\orb}{PointRubric}
\newcommand{\openrubrics}{OpenRubrics}
\newcommand{\rar}{RaR-Science}
\newcommand{\gpqa}{GPQA-Diamond}
\newcommand{\gptfour}{GPT-4o}

\newcommand{\rqone}{Can 1.7B models serve as criterion-level rubric judges, and which readout works best?}
\newcommand{\rqtwo}{Can the selected probe judge be used as an RL reward model?}
\newcommand{\rqthree}{Does the learned signal transfer beyond the RaR-Science training setting?}
\newcommand{\rqfour}{What is the efficiency-quality tradeoff compared with a Generative judge?}

\title{Small Language Models as Judges for Rubric-Based \\ Reinforcement Learning}

\author{
  Fengyu Xie$^{1}$ \quad
  Yilun Zhao$^{2}$ \quad
  Bingsen Chen$^{1}$ \quad
  Arman Cohan$^{2}$ \quad
  Chen Zhao$^{1}$ \\[0.45em]
  $^{1}$New York University \qquad $^{2}$Yale University \\[0.25em]
  \texttt{\{fx2137,chen.zhao\}@nyu.edu} \quad
  \texttt{yilun.zhao@yale.edu}
}
\begin{document}
\maketitle

\begin{abstract}

Rubric-based reinforcement learning extends RL beyond tasks with exact answers or rule-based verifiers by scoring responses against instance-specific criteria.
However, this makes reward computation expensive: training requires repeated rubric judging, often with proprietary APIs or local generative LLM judges with 7B parameters or more. We study whether smaller language models can serve as efficient and reliable rubric-based judges. To make this question measurable, we construct \orb{} and \rar{}-Static, two pointwise rubric-based evaluation datasets with instance-specific criteria and itemwise satisfaction labels. We compare three ways of extracting criterion-level judgments from small models: Generative verdicts, Yes/No Logprob scoring, and Probe judges. Across both datasets, the Qwen3-1.7B Probe judge achieves the strongest criterion-level agreement among these methods, outperforming Generative and Logprob judges. Used as a GRPO reward model, it trains a policy from 0.232 to 0.643 on \rar{} rubric score, compared with 0.594 for an 8B Generative judge baseline, while the baseline requires 10.7$\times$ more reward-judge time. Task and domain transfer experiments further suggest that Probe judges preserve criterion-level reward structure across settings. \footnote{Our code and data have been released at \url{https://github.com/Ignotus6043/SLM-rubric-RL}.}
 


\end{abstract}

\input{Sections/1-Intro}
\input{Sections/3-Dataset}
\input{Sections/4-Method}

\input{Sections/5-Experiment}

\input{Sections/6-RubricRL}
\input{Sections/2-Related}

\section{Conclusion}

We study whether small LMs can serve as criterion-level rubric judges for rubric-based RL. We derive \orb{} from \openrubrics{} and construct \rar{}-Static from \rar{} to test the evaluation capabilities of small judges.
 Across both datasets, Probe judges perform best, showing that 1.7B-scale models encode useful rubric-satisfaction signals in hidden states.
When used as an RL reward model, the Qwen3-1.7B Probe judge trains a policy that outperforms an 8B Generative judge reward baseline while using far less judge time. Transfer experiments evaluate both policy transfer and judge transfer: the Probe-reward policy improves GPQA-Diamond accuracy, while a RaR-Science-trained Probe retains criterion-level agreement on RaR-Medicine. Overall, small Probe judges provide an effective and efficient reward model for rubric-based RL.


\section*{Limitations}

Our supervised target is \gptfour{} criterion scoring, not direct human ground
truth. This matches the operational goal of the paper: approximating the
expensive rubric judge used in the studied rubric-RL pipeline. Human preference
comparisons and independent rescoring in
Appendix~\ref{app:additional-results-tables} support the validity of this
target in our setting, but broader human evaluation would be needed to
characterize agreement across tasks and domains.

Our strongest evidence is on \rar{} questions, with additional checks
on \gpqa{} and RaR-Medicine. These results show useful transfer beyond the
calibration setting, but do not imply domain-invariant rubric judgment. New
domains, especially dialogue, long-form instruction following, or
safety-critical expert settings, may require additional calibration and
validation. Similarly, the RL experiments control the actor, reward aggregation,
and training configuration, but cover one policy family and one rubric-RL setup;
we therefore treat the human audit and independent-reference checks as
setting-specific evidence against reward overoptimization.

Finally, the efficiency results are measurements of our implementation and
hardware. We report cumulative judge time, validation reward time, and elapsed RL time from matched runs. The judge-time advantage is larger than the wall-clock advantage because reward calls are parallelized and other RL costs remain in the loop, so the reported ratios should be interpreted as empirical measurements for this setup rather than universal serving constants.

\section*{Ethical considerations}

This work reuses two public research datasets, \openrubrics{} and \rar{} \citep{openrubrics, rar}. We use these resources for research on rubric judging and cite the original papers. \orb{} is derived from \openrubrics{}, and \rar{}-Static is derived from \rar{} by adding a fixed response bank and reference labels for static judge evaluation. We do not claim ownership of the original prompts, rubrics, or reference answers. Any released version of our derived data should retain the original source attribution and follow the usage terms provided by the original dataset authors.

The datasets contain model prompts, rubrics, reference answers, and model responses. We do not collect or add private user data, demographic attributes, or personal identifiers. During construction and validation, we manually inspected sampled examples and generated responses for obvious personally identifying information and unsafe or offensive content. We did not intentionally retain any newly introduced personal identifiers. Because \orb{} and \rar{}-Static are derived from public research datasets and model-generated responses, residual unsafe, biased, sensitive, or offensive content may still be inherited from the source artifacts or produced by models. We therefore treat the derived artifacts as research-only evaluator resources, and our benchmark should not be treated as a guarantee that a judge is safe for deployment.

We conduct two small human audits: one audit of \orb{} reference labels and one blind pairwise audit of the main RL comparison. Annotators are asked only to evaluate model responses against provided rubrics or to compare two model responses. The annotators were members of the research team, so no external recruitment or payment was involved. The tasks do not ask annotators to disclose personal information, and we store and report only aggregate agreement or preference statistics. The audits are used as reliability checks for the operational \gptfour{} reference target, not as standalone human preference datasets or as a replacement for broader human evaluation.

\section*{Acknowledgements}

The authors gratefully acknowledge the NYU High Performance Computing Torch cluster for providing the computational resources used in this work.

\bibliography{acl_anthology,custom}

\input{Sections/Appendix}

\end{document}

%% file: Sections/1-Intro.tex
\section{Introduction}

\begin{figure*}[t]
\centering
\includegraphics[width=0.98\textwidth,trim=16pt 14pt 16pt 14pt,clip]{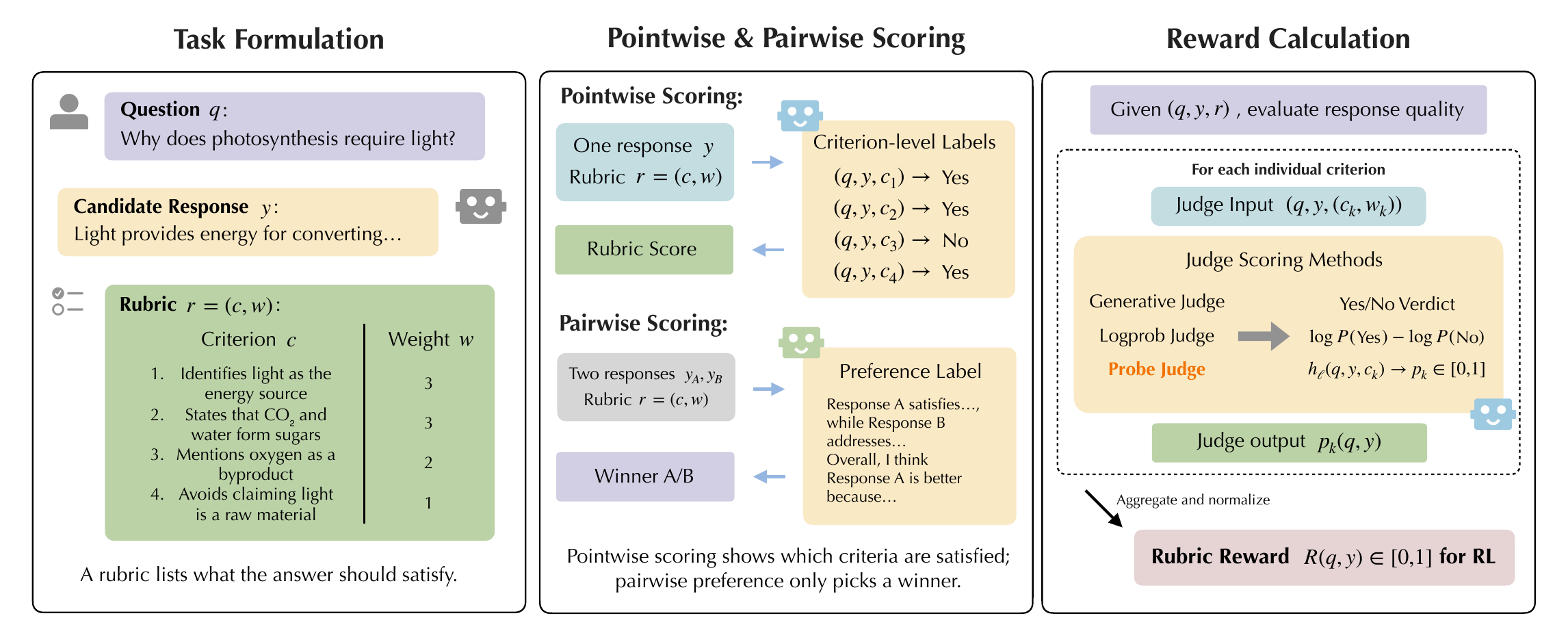}
\caption{Rubric scoring setup. A rubric specifies criteria and weights for a candidate response. Pointwise scoring yields criterion-level labels, while pairwise scoring yields a relative winner. Criterion scores can then be aggregated into a scalar reward for evaluation or reinforcement learning.}
\label{fig:rubric-scoring-overview}
\end{figure*}

Recent progress in reinforcement learning with verifiable rewards (RLVR) has shown that outcome rewards from rule-based verifiers can effectively elicit reasoning abilities in large language models (LLMs) on verifiable tasks such as math and coding \citep{code-rl, rltf, grpo, rlvr, deepseek-r1}. However, many increasingly important agentic applications go beyond this narrow notion of verifiability. In long-form generation tasks such as Deep Research, quality can depend on content coverage, factuality, source use, and task-specific constraints, which are hard to capture with simple automatic checkers \citep{gemini-deep-research, healthbench, openai-deep-research, dr-tulu}. Rubric-based reinforcement learning offers a promising alternative for such settings: rubrics evaluate multiple aspects of generation quality through instance-specific criteria while still producing scalar rewards for optimization \citep{rlhf, rar}.

A key component of rubric-based RL is obtaining the rubric reward, where an LLM judge evaluates responses against explicit rubric criteria and aggregates these judgments into rewards for RL training \citep{llm-as-judge, kim-etal-2024-prometheus}. Existing rubric-based RL approaches mainly rely on proprietary model APIs or large generative judges to maintain judgment quality \citep{gpt4o, rar, dr-tulu}. This makes reward computation expensive at scale because each RL step may require scoring long responses against multiple criteria. Proprietary judges also make exact reproduction harder because access depends on a fixed model snapshot from the provider. These limitations raise a central question: \emph{Can much smaller language models serve as effective rubric judges for rubric-based RL?}

To answer this question, we need evaluation data that matches the reward computation used in rubric-based RL: explicit criteria, multiple responses under the same rubric, labels for each response--criterion pair, and a weighted scoring rule. Existing resources provide important pieces of this setup, but not the full structure required for this evaluation setting. Preference and reward-model datasets focus on pairwise comparisons or scalar reward modeling \citep{rlhf, lambert-etal-2025-rewardbench, judgelm}.
Fine-grained and rubric-evaluation benchmarks provide skills, criteria, or judge-evaluation targets, but typically lack weighted rubric rewards or response--criterion satisfaction labels for fixed candidate responses \citep{flask, openrubrics, rubricbench, rubriceval, if-rewardbench}.
Domain-specific rubric datasets provide structured rubrics and scoring rules, but not controlled static response banks with labels for every response--criterion pair \citep{healthbench, rar}. We therefore derive \orb{}, a controlled pointwise benchmark from \openrubrics{}, and construct \rar{}-Static, a fixed response bank of similar format based on \rar{}, for evaluating the same judge methods under the rubric-based RL format.

We compare three rubric-judging methods on \orb{} and \rar{}-Static using Qwen3 models from 0.6B to 8B parameters \citep{qwen3}: Generative verdicts, Yes/No Logprob scoring, and hidden-state Probes. For each Probe, the language-model backbone is frozen and only a lightweight linear classifier is trained on GPT-4o criterion labels. Across both static settings, Probe judges substantially outperform Generative and Logprob judges, suggesting that even 1.7B-scale models contain useful evaluative signals in their hidden representations that Generative and Logprob scoring do not recover as reliably \citep{bert-as-a-judge, representation-judge}.

We then use the Qwen3-1.7B Probe judge as the reward model for GRPO in a controlled \rar{} science setup \citep{grpo, rar}. This Probe reward improves the policy's rubric score from 0.232 to 0.643, outperforming an 8B Generative judge reward baseline, which reaches 0.594, while the 8B baseline requires 10.7$\times$ more reward-judge time at the comparison checkpoint. Additional transfer experiments on \gpqa{} and cross-domain rubric splits show that the Probe signal is not limited to the static training setting \citep{gpqa}.

\noindent Our contributions are as follows:
\begin{itemize}[noitemsep,topsep=0pt,leftmargin=*]
    \item We construct two pointwise rubric-evaluation datasets for studying small rubric judges: \orb{}, adapted from \openrubrics{}, and \rar{}-Static, built from \rar{} to match the rubric format in downstream RL.
    \item We compare Generative, Logprob, and Probe judges on \orb{} and \rar{}-Static. Probe judges perform best, showing that small models encode useful criterion-level evaluative signals that are more reliably exposed through hidden states than through generation.
    \item We use the Qwen3-1.7B Probe judge as the GRPO reward model on \rar{}, training a policy from 0.232 to 0.643 on \rar{} rubric score, compared with 0.594 for an 8B Generative judge reward baseline; the 8B Generative baseline requires 10.7$\times$ as much cumulative reward-judge time.
\end{itemize}

%% file: Sections/3-Dataset.tex
\section{Rubric Judging Task and Dataset}

\input{tables/benchmark_gap}
\label{sec:task-data}

A key component of rubric-based RL is rubric judging: determining whether a model response satisfies a given rubric criterion. We first define the task of pointwise rubric judging, where the judge evaluates one response against one rubric item at a time (\S\ref{sec:pointwise-pairwise}). We then present the dataset design, review why existing datasets are not directly suitable for this setting, and describe how we adapt existing resources into pointwise rubric-judging benchmarks (\S\ref{sec:data-construction}--\S\ref{sec:rar-construction}).

\subsection{Our Task: Pointwise Rubric Judging}
\label{sec:pointwise-pairwise}

Preference-based alignment methods \citep{rlhf,llm-as-judge,lambert-etal-2025-rewardbench} typically learn reward signals from pairwise response comparisons: given a prompt and two candidate responses, a human annotator, reward model, or LLM judge predicts which response better satisfies the user's intent. This formulation is useful for learning relative preferences, but it collapses all evaluation criteria into a single comparison. As a result, it does not reveal which aspects of a response are correct or incorrect.

Rubric-based RL instead specifies an explicit rubric $C(q)=\{(c_j,w_j)\}_{j=1}^{m_q}$, where each $c_j$ is a natural-language criterion and $w_j$ is its aggregation weight. We therefore study \emph{pointwise rubric judging}: given a prompt $q$, a candidate response $y$, and one rubric item $c_j$, the judge takes $(q,y,c_j)$ as input and predicts whether response $y$ satisfies criterion $c_j$. The judge may return either a binary satisfaction label or a soft satisfaction score. These item-level outputs are then aggregated with the rubric weights to produce the scalar rubric score used for evaluation or RL.

Pointwise rubric judging can still induce pairwise preferences by comparing the aggregated scores of two responses under the same prompt and rubric. However, unlike pairwise judging, it preserves the rubric structure: it identifies which criteria are satisfied or violated and supports direct scalar reward computation. Appendix~\ref{app:rubric-scoring} gives additional formal details and examples.

\subsection{Dataset Design Overview}
\label{sec:data-construction}

Existing rubric and reward-model datasets provide useful supervision for judging model outputs. However, they are not directly designed for pointwise rubric judging, where the judge must determine whether a single response satisfies a single rubric criterion and support reward computation from these item-level decisions.

As shown in \autoref{tab:benchmark-gap}, our setting requires four components: explicit rubric criteria as judge targets, multiple responses under the same prompt and rubric, per-criterion satisfaction labels, and a weighted scoring rule for aggregating criterion-level judgments into scalar rewards. Existing resources cover parts of this design, but usually lack criterion-level labels, controlled response banks, or the weighted aggregation structure used in rubric-based RL. We therefore use two evaluation settings: \orb{}, a new pointwise benchmark derived from \openrubrics{}, and \rar{}-Static, a fixed response bank built from \rar{} for static judge evaluation.

\subsection{\orb{} Construction}
\label{sec:orb-construction}

\orb{} evaluates criterion-level rubric judging. It is derived from \openrubrics{}, which provides general-domain prompts and detailed natural-language rubrics. Since the original supervision is pairwise and does not label criterion-level satisfaction, we add new candidate responses and per-criterion labels to convert \openrubrics{} into a controlled pointwise benchmark.

\paragraph{Rubric standardization.}

We sample 3,000 \openrubrics{} questions with explicit hard-rule constraints. We rewrite each one into six criteria: two hard criteria for mandatory constraints and four soft criteria for response quality. This fixed format gives all prompts the same criterion and weight structure while preserving the distinction between required constraints and softer quality dimensions. A manual audit found that 47 of 50 sampled transformations preserved the original rubric's core intent; further results and examples are provided in Appendix~\ref{app:orb-transformation-audit}.

\paragraph{Response roles.}
For each prompt, we generate four candidate responses under the same rubric: a full-score response, a low-score response, a rubric-guided partial response, and an unguided response. These roles are not treated as four ordinal quality bins. Instead, the full-score and low-score responses provide fixed anchors, while the partial and unguided responses introduce broader intermediate variation under the same prompt and rubric. In the final benchmark, full-score responses all receive score 10, low-score responses all fall in the 0--3 score range, and the partial and unguided responses cover broader score distributions. We then use \gptfour{} \citep{gpt4o} as the operational reference judge to assign a binary satisfaction label to every response--criterion pair.

\paragraph{Final benchmark.}
After filtering and validation, \orb{} contains 1,042 questions, 4,168 responses, and 25,008 response--criterion labels. For evaluation, we use a question-level split of 417 training questions, 104 development questions, and 521 held-out questions, so all responses to the same prompt remain within a single split. Appendices~\ref{app:orb-construction-details}--\ref{app:orb-reference-judge} give construction details; Appendix~\ref{app:dataset-example} gives a concrete example for \orb.

\subsection{\rar{}-Static}
\label{sec:rar-construction}

\rar{}-Static complements \orb{} by testing the same pointwise judge methods in the downstream science-domain rubric format. It is built from \rar{}, which uses instance-specific rubrics with variable numbers of criteria and absolute criterion weights \citep{rar}. Unlike the RL experiments in \S\ref{sec:rl}, \rar{}-Static is used only for static judge evaluation.

\paragraph{Response bank and split.}
\rar{}-Static contains 1,500 randomly sampled science questions. Each question keeps the original \rar{} reference answer and receives one additional candidate response generated by greedy decoding from Qwen3-4B \citep{qwen3}, giving 3,000 candidate responses evaluated under the same rubric. The rubrics contain between 6 and 11 criteria, with an average of 7.52 criteria per question. For evaluation, we use a fixed split of 1,000 training questions, 200 development questions, and 300 test questions.

\paragraph{Reference labels and aggregation.}
We apply the same pointwise labeling protocol used for \orb{}: \gptfour{} assigns a binary satisfaction label to each response--criterion pair. Scalar scores use the original \rar{} criterion weights after converting pitfall criteria into the same satisfaction-label interface. Specifically, pitfall criteria with negative raw weights are treated as avoidance criteria, and scalar scores are aggregated with absolute weights so that higher satisfaction always corresponds to better rubric compliance. Appendices~\ref{app:rar-static-bank} and \ref{app:reference-aggregation} give the response-bank construction and aggregation details; Appendix~\ref{app:dataset-example} also contains a concrete example of \rar{}-Static.

\subsection{Reference Labels and Human Audit}
\label{sec:data-validation}

Both data settings use \gptfour{} as the operational reference judge for per-criterion labels. We treat these labels as an operational reference target rather than as human ground truth. The purpose of the benchmark is to test whether smaller local judges can approximate an expensive rubric-grading pipeline, so the human audit is a reliability check on this target rather than a replacement for larger human evaluation.

We manually audit 100 sampled responses from the held-out \orb{} split, covering 600 decisions and 50 paired response comparisons. The audit achieves 90.9\% weighted criterion agreement with \gptfour{} and 96.0\% pairwise preference agreement. Disagreements are concentrated in softer qualitative criteria, such as completeness, specificity, and clarity, rather than hard factual or constraint-following criteria. Appendix~\ref{app:dataset-validation} gives the audit protocol and additional examples.

%% file: tables/benchmark_gap.tex
\newcommand{\benchyes}{\textcolor{green!50!black}{\(\checkmark\)}}
\newcommand{\benchno}{\textcolor{red!70!black}{\(\times\)}}

\begin{table*}[t]
\centering
\small
\resizebox{\linewidth}{!}{%
\setlength{\tabcolsep}{5pt}
\begin{tabular}{lcccc}
\toprule
\textbf{Resource} & \textbf{Rubric criteria} & \textbf{Multiple responses} & \textbf{Per-criterion labels} & \textbf{Weighted rubric scores} \\
\midrule
FLASK \citep{flask} & \benchyes & \benchyes & \benchno & \benchno \\
HealthBench \citep{healthbench} & \benchyes & \benchno & \benchno & \benchyes \\
IF-RewardBench \citep{if-rewardbench} & \benchyes & \benchyes & \benchyes & \benchno \\
JudgeLM-100K \citep{judgelm} & \benchno & \benchyes & \benchno & \benchno \\
\openrubrics{} \citep{openrubrics} & \benchyes & \benchyes & \benchno & \benchno \\
\rar{} / RaR-Medicine \citep{rar} & \benchyes & \benchno & \benchno & \benchyes \\
RewardBench \citep{lambert-etal-2025-rewardbench} & \benchno & \benchyes & \benchno & \benchno \\
RubricBench \citep{rubricbench} & \benchyes & \benchyes & \benchno & \benchno \\
RubricEval \citep{rubriceval} & \benchyes & \benchno & \benchyes & \benchno \\

\midrule

\textbf{\orb{} (ours)} & \benchyes & \benchyes & \benchyes & \benchyes \\
\textbf{\rar{}-Static (ours)} & \benchyes & \benchyes & \benchyes & \benchyes \\
\bottomrule
\end{tabular}
}
\caption{Data requirements for pointwise rubric-based judge evaluation. \orb{} and \rar{}-Static provide explicit rubric criteria, multiple responses per prompt, per-criterion labels, and weighted rubric scores.} 
\label{tab:benchmark-gap}
\end{table*}

%% file: Sections/4-Method.tex
\section{Rubric-Judge Methods}
\label{sec:methods}

\looseness=-1
We now describe how each small language model is used as a rubric judge. All methods receive the same item-level input \((q,y,c_j)\): a prompt, a candidate response, and one rubric criterion. Each method predicts whether the response satisfies the criterion, either as a binary verdict or as a satisfaction probability. The resulting criterion-level outputs are aggregated with the same rubric weights. We compare three scoring methods: Generative, Logprob, and Probe. We also use SFT as a post-training test, then apply the same three methods to the SFT backbone. \autoref{tab:method-matrix} summarizes the method matrix. Complete prompt templates used throughout the study are provided in Appendix~\ref{app:prompts}.

\input{tables/method_matrix}

\subsection{Scoring Methods}
\label{sec:judge-scoring-methods}

We compare three ways to extract criterion-level satisfaction signals from the same language-model backbone: generating a verdict, scoring fixed verbalizers, and probing hidden states.

\paragraph{Generative Judges.}
Generative judges prompt the language model to produce a binary satisfaction verdict for one criterion. Parsed verdicts are aggregated with the rubric weights. Outputs that cannot be parsed into the required binary format are treated as invalid predictions.

\paragraph{Logprob Judges.}
Logprob judges replace verdict generation with forced-choice scoring. Given the same item-level input, the model is asked to score the next-token probabilities of the fixed verbalizers ``Yes'' and ``No'' under the criterion prompt and compute their log-probability margin:
\[
    \Delta_j
    =
    \log P(\text{Yes}\mid q,y,c_j)
    -
    \log P(\text{No}\mid q,y,c_j).
\]
Binary static metrics threshold this margin at 0. Scalar rubric aggregation uses the soft score \(p_j(q,y)=\sigma(\Delta_j)\). This method avoids parsing generated text. Appendix~\ref{app:generated-logprob-details} gives the verbalizers and scoring details.

\paragraph{Probe Judges.}
Probe judges test whether hidden states encode the rubric judgment even when the model may not express it reliably through generated text \citep{representation-judge}. We render the same single-criterion prompt, keep all language-model parameters frozen, and extract the final non-padding-token hidden state \(h_\ell(q,y,c_j)\) at layer \(\ell\). Only a lightweight linear classifier head is trained, using binary cross-entropy to predict the \gptfour{} criterion label from this representation. Layer selection and the binary threshold use the development split. Scalar rubric scores and RL rewards use the unthresholded satisfaction probability. Appendix~\ref{app:probe-details} gives the training and calibration details.

\subsection{Post-training with Supervised Fine-tuning}
\label{sec:sft-backbones}

Supervised fine-tuning (SFT) is our post-training test: it asks whether training a backbone to predict rubric verdicts makes it a better rubric judge. Each training example contains a prompt, a candidate response, the complete rubric, and the corresponding vector of \gptfour{} Yes/No verdicts, following standard supervised instruction tuning \citep{rlhf}. After SFT, we evaluate the resulting backbone with the same item-level Generative, Logprob, and Probe methods used for the base model. Therefore, SFT Generative, SFT Logprob, and SFT Probe differ from the corresponding base rows only in the backbone parameters, not in the scoring procedure. Appendix~\ref{app:sft-details} gives the SFT data format and evaluation setup.

\input{tables/static_judge_results}

%% file: tables/method_matrix.tex
\begin{table}[t]
\centering
\footnotesize
\setlength{\tabcolsep}{2pt}
\resizebox{\columnwidth}{!}{%
\begin{tabular}{@{}lcclc@{}}
\toprule
\textbf{Method} & \textbf{LM tuned?} & \textbf{Head?} & \textbf{Output} & \textbf{RL use} \\
\midrule
Generative & \benchno & \benchno & Binary verdict & Baseline \\
Logprob & \benchno & \benchno & Satisfaction probability & -- \\
Probe & \benchno & \benchyes & Satisfaction probability & Main \\
SFT Generative & \benchyes & \benchno & Binary verdict & -- \\
SFT Logprob & \benchyes & \benchno & Satisfaction probability & -- \\
SFT Probe & \benchyes & \benchyes & Satisfaction probability & -- \\
\bottomrule
\end{tabular}%
}
\caption{Rubric-judge method matrix. All methods use the same item-level input $(q,y,c_j)$ and the same rubric-weight aggregation rule. SFT changes the backbone state, while Generative, Logprob, and Probe define the criterion-level readout.}
\label{tab:method-matrix}
\end{table}

%% file: tables/static_judge_results.tex
\begin{table*}[!t]
\centering
\scriptsize
\setlength{\tabcolsep}{2pt}
\resizebox{\linewidth}{!}{%
\begin{tabular}{@{}lcccccc@{\hspace{12pt}}cccccc@{}}
\toprule
\multicolumn{1}{c}{}
& \multicolumn{6}{c}{\textbf{\orb{}}}
& \multicolumn{6}{c}{\textbf{\rar{}-Static}} \\
\cmidrule(lr){2-7}\cmidrule(lr){8-13}
& \multicolumn{3}{c}{\textbf{Base model}} & \multicolumn{3}{c}{\textbf{SFT model}}
& \multicolumn{3}{c}{\textbf{Base model}} & \multicolumn{3}{c}{\textbf{SFT model}} \\
\cmidrule(lr){2-4}\cmidrule(lr){5-7}
\cmidrule(lr){8-10}\cmidrule(lr){11-13}
\textbf{Model} & \textbf{Generative} & \textbf{Logprob} & \textbf{Probe}
& \textbf{Generative} & \textbf{Logprob} & \textbf{Probe}
& \textbf{Generative} & \textbf{Logprob} & \textbf{Probe}
& \textbf{Generative} & \textbf{Logprob} & \textbf{Probe} \\
\midrule
Qwen3-0.6B
& 0.370 & 0.582 & 0.802
& 0.560 & 0.604 & \textbf{0.820}
& 0.413 & 0.433 & 0.793
& 0.607 & 0.605 & \textbf{0.796} \\
Qwen3-1.7B
& 0.518 & 0.766 & 0.875
& \textbf{0.902} & 0.713 & 0.845
& 0.443 & 0.449 & \textbf{0.835}
& 0.708 & 0.520 & 0.828 \\
Qwen3-4B
& 0.790 & 0.893 & \textbf{0.913}
& 0.808 & 0.892 & 0.906
& 0.496 & 0.738 & \textbf{0.851}
& 0.673 & 0.794 & 0.845 \\
Qwen3-8B
& 0.888 & 0.907 & \textbf{0.914}
& 0.846 & 0.895 & 0.912
& 0.609 & 0.756 & \textbf{0.864}
& 0.642 & 0.754 & 0.861 \\
\bottomrule
\end{tabular}
}
\caption{Static evaluator results on held-out splits. \orb{} values are weighted criterion accuracy with four candidate responses per question; \rar{} values are criterion-level macro-F1 against \gptfour{} labels with two candidate responses per question. Bold marks the best method for each model size within each benchmark.}
\label{tab:static-final}
\end{table*}

%% file: Sections/5-Experiment.tex
\section{Static Rubric-Judge Evaluation}
\label{sec:static}

\begin{questionbox}
\textcolor{orange!50!yellow}{\faLightbulb} \xspace
\textbf{Q1:} \rqone
\end{questionbox}

\noindent Before using a rubric judge as an RL reward model, we first evaluate whether small LMs can approximate criterion-level judgments on fixed response banks.
Specifically, each judge receives a question, candidate response, and rubric criterion, and predicts whether the response satisfies that criterion.
We compare three readouts from the same Qwen3 backbones: Generative verdicts, Yes/No Logprob scoring, and hidden-state Probes.
All methods are evaluated against \gptfour{} \citep{gpt4o} criterion labels as the operational reference.

\subsection{Experiment Setup}
\label{sec:static-setup}

We evaluate Qwen3 judge backbones at four scales: 0.6B, 1.7B, 4B, and 8B parameters \citep{qwen3}.
For each backbone, we compare Generative, Logprob, and Probe judges, using both base and SFT models when applicable.
All methods are evaluated on the same held-out question splits, and trained or calibrated components use only the corresponding training and development data.
Appendix~\ref{app:static-targets} defines the static criterion-satisfaction target and explains how binary verdicts, probabilities, and scalar rubric scores are computed.

We report one primary metric for each data setting.
On \orb{}, the headline metric is weighted criterion accuracy, matching its hard/soft criterion weighting.
On \rar{}-Static, the headline metric is criterion-level macro-F1, since the downstream rubric setting has variable rubric lengths, weights, and label balance.
Complementary metrics are reported in Appendix~\ref{app:static-metrics}.

\subsection{Static Judge Results}
\label{sec:static-results}

Table~\ref{tab:static-final} compares three ways of extracting criterion-level judgments from the same Qwen3 backbones.

\paragraph{\orb{} shows that small models can perform pointwise rubric judging when the benchmark format is controlled.} Base generative judge improves with scale, from 0.370 weighted criterion accuracy at 0.6B to 0.888 at 8B.
SFT is especially effective for the 1.7B generative judge, which reaches 0.902, showing that a small model can learn the explicit Yes/No criterion-verdict format in this controlled setting.

\paragraph{On \rar{}-Static, the readout method matters more than model size alone.} For the same 1.7B base backbone, Generative and Logprob reach only 0.443 and 0.449 macro-F1, while Probe reaches 0.835.
This suggests that useful rubric-satisfaction information is present in the model's hidden states but is not recovered reliably by Generative or Logprob scoring. An item-level comparison supports this interpretation: among 4,518 held-out criterion decisions, 33.0\% are correct only under Probe, whereas 7.8\% are correct only under Generative. Appendix~\ref{app:static-readout-errors} reports the complete error decomposition and representative examples of both error directions.

\paragraph{Probe is the strongest readout on \rar{}-Static across all model sizes.}
The 1.7B Probe reaches 0.835 macro-F1, compared with 0.864 for the 8B Probe, while remaining substantially smaller than the 8B Generative judge baseline used later in RL.
This makes Qwen3-1.7B Probe the natural candidate for the RL reward experiments: it is much smaller than the 8B Generative judge baseline while retaining strong criterion-level agreement on the downstream rubric format.
Under response-generator shift, the unchanged 1.7B Probe reaches 0.741 macro-F1 on an extended 1,200-response bank with Mistral-7B-Instruct-v0.3 and OLMo-2-1124-7B-Instruct responses \citep{mistral7b,olmo2}, versus 0.394 for Generative and 0.424 for Logprob.
Appendix~\ref{app:rar-extended-bank} reports the protocol and complementary metrics; Appendix~\ref{app:bootstrap-ci} gives confidence intervals for Table~\ref{tab:static-final}.

\noindent\textbf{Comparison with a pairwise rubric reward model.}
Rubric-RM is designed to predict preferences between response pairs rather than criterion-level scores for individual responses \citep{openrubrics}. We therefore evaluate it as a static compatibility baseline on \gptfour{} non-tied pairs, using the scalar scores from our pointwise judges to induce comparable pairwise rankings. The Qwen3-1.7B Probe reaches pairwise accuracies of 0.924 on \orb{} and 0.888 on \rar{}-Static. Across Rubric-RM-4B and Rubric-RM-8B, the best corresponding accuracies are 0.322 and 0.487 when parse failures are counted as incorrect, or 0.575 and 0.670 when evaluation is restricted to parseable outputs. These results show that Rubric-RM captures useful pairwise signal when its outputs are parseable, but its pairwise interface and frequent formatting failures prevent it from serving as a direct criterion-level reward in our RL pipeline. Appendix~\ref{app:rubric-rm-pairwise} reports the per-model results, parse rates, and evaluation conventions.

\subsection{Selecting the Probe Configuration for RL}
\label{sec:probe-selection}

Table~\ref{tab:probe-ablation-final} studies the data efficiency and readout design of Qwen3-1.7B Probes on \rar{}-Static.
The linear last-token probe is already strong with limited supervision, reaching 0.805 macro-F1 with 100 training questions and 0.834 with 1,000 training questions.
The architecture variants provide only modest gains: MLP heads and last-4-layer averaging are close to the linear last-token probe, while mean pooling is substantially worse.
We use the same linear last-token design for the main RL reward, fitted separately on 450 training and 50 development responses as described in Appendix~\ref{app:probe-details}.
It is simpler, nearly matches the strongest ablations, directly produces criterion-level probabilities, and avoids generated-verdict parsing. The full layer sweep in Appendix~\ref{app:probe-layer-selection} shows that layer selection lies on a broad late-layer plateau rather than a single isolated optimum. This configuration gives a compact reward judge, so Section~\ref{sec:rl} evaluates whether it remains useful inside the RL loop as an online reward model.

\input{tables/probe_configuration}

%% file: tables/probe_configuration.tex
\begin{table}[t]
\centering
\footnotesize
\setlength{\tabcolsep}{5pt}
\resizebox{\linewidth}{!}{%
\begin{tabular}{@{}clrc@{}}
\toprule
\textbf{Ablation} & \textbf{Probe setting} & \textbf{\# Train q.} & \textbf{Macro-F1} \\
\midrule
 & Linear, last token & 50 & 0.767 \\
 & Linear, last token & 100 & 0.805 \\
Data & Linear, last token & 250 & 0.818 \\
 & Linear, last token & 500 & 0.828 \\
 & Linear, last token & 1000 & 0.834 \\
\midrule
 & Linear, last token & 1000 & 0.834 \\
 & MLP-32, last token & 1000 & 0.834 \\
Design & MLP-64, last token & 1000 & \textbf{0.840} \\
 & Linear, mean pooling & 1000 & 0.761 \\
 & Linear, last-4-layer mean & 1000 & 0.839 \\
\bottomrule
\end{tabular}
}
\caption{Qwen3-1.7B Probe ablations on \rar{}-Static. Data rows vary the number of training questions for the linear last-token probe. Design rows use 1,000 training questions and compare alternative heads and pooling strategies.}

\label{tab:probe-ablation-final}
\end{table}

%% file: Sections/6-RubricRL.tex
\section{Probe-Based Rubric Rewards for RL}
\label{sec:rl}

\noindent Section~\ref{sec:static} selects Qwen3-1.7B Probe as a compact criterion-level judge.
We now test whether this judge remains useful when used inside the RL loop, whether the learned signal transfers beyond the exact training setting, and how its quality-efficiency tradeoff compares with a larger generative judge.

\subsection{Matched RL Protocol and Results}
\label{sec:rl-main}

\begin{questionbox}
\textcolor{orange!50!yellow}{\faLightbulb} \xspace
\textbf{Q2:} \rqtwo
\end{questionbox}

\input{tables/matched_rl_results}

\input{tables/gpqa_transfer}

\input{tables/cross_domain_transfer}

\input{tables/efficiency_quality}

\noindent We test whether the selected Qwen3-1.7B Probe can serve as the reward model for rubric-based RL.
All controlled runs train the same Qwen3-4B-Base actor with GRPO \citep{grpo} on the same \rar{} training set. The actor, optimizer, schedule, response budget, aggregation rule, and final \gptfour{} evaluator are fixed; only the reward judge changes. Full configuration details are given in Appendix~\ref{app:rl-implementation-details}, and development checks are summarized in Appendix~\ref{app:rl-configuration-checks}.

\paragraph{The Qwen3-1.7B Probe reward produces the strongest trained policy.}
Table~\ref{tab:rl-final} reports the controlled reward-model comparison. Starting from the same base actor, the Qwen3-1.7B Probe reward improves the final \rar{} rubric score from 0.232 to 0.643. Under the same training and evaluation protocol, the larger Qwen3-8B Generative reward reaches 0.594. Thus, the compact Probe reward trains a higher-scoring policy than the larger Generative judge in this matched RL setting. Appendix~\ref{app:rl-score-ci} reports bootstrap confidence intervals, and Appendix~\ref{app:reward-dynamics} gives representative policy improvements and failure cases.

\paragraph{Downstream reward utility does not exactly follow static judge accuracy.}
Although the 4B and 8B Probes achieve slightly higher fixed-bank macro-F1 in Table~\ref{tab:static-final}, the 1.7B Probe produces the strongest externally evaluated policy after GRPO. On-policy reward diagnostics suggest a possible explanation: the larger Probes assign more saturated rewards to actor rollouts, reducing the within-group reward variation available to GRPO. We therefore select the reward judge by downstream policy improvement under the controlled RL protocol. Appendix~\ref{app:reward-dynamics} reports the corresponding reward-distribution and saturation statistics.

\paragraph{A blind human audit supports the main policy comparison.}
The probe-reward policy is preferred over the base actor on 72 out of 100 examples, while \gptfour{} rubric preference favors the probe policy on 80 out of the same 100 pairs. After excluding human Tie/Unsure labels and \gptfour{} exact ties, human and \gptfour{} preferences agree on 69 of 72 cases (95.8\%). Among the 13 human ties, \gptfour{} also assigns similar scores: 7 have an absolute score gap at most 0.05, and 11 have a gap below 0.20. These checks support the operational \gptfour{} target rather than replacing a larger human evaluation. Appendix~\ref{app:human-audit} gives the audit protocol, tie analysis, and reference-judge validation.

\subsection{Transfer Results}
\label{sec:generalization}

\begin{questionbox}
\textcolor{orange!50!yellow}{\faLightbulb} \xspace
\textbf{Q3:} \rqthree
\end{questionbox}

\noindent We evaluate transfer in two ways: whether the policy trained with the probe reward improves outside the \rar{} rubric format, and whether the Qwen3-1.7B Probe's criterion-satisfaction signal transfers from science rubrics to medical rubrics without target-domain fitting.
The evaluation protocols are defined in Appendix~\ref{app:generalization-protocols}.

Table~\ref{tab:gpqa-transfer-final} compares the base Qwen3-4B actor with the policy trained using the Qwen3-1.7B Probe reward.
On 198 \gpqa{} questions \citep{gpqa} across four answer-order runs, the probe-reward policy improves accuracy from 0.335 to 0.388, suggesting that \textbf{the policy improvement can generalize beyond the \rar{} rubric-scoring format.}

Table~\ref{tab:cross-domain-final} evaluates whether the probe's criterion-satisfaction signal transfers across domains.
A Qwen3-1.7B linear probe trained on \rar{} criterion labels reaches 0.718 macro-F1 on RaR-Medicine without Medicine labels for fitting.
This is below the Medicine-trained in-domain probe, but shows that the \textbf{RaR-Science-trained probe retains meaningful criterion-level agreement under a different rubric distribution.}

\subsection{Efficiency-Quality Results}
\label{sec:efficiency}

\begin{questionbox}
\textcolor{orange!50!yellow}{\faLightbulb} \xspace
\textbf{Q4:} \rqfour
\end{questionbox}

\noindent Table~\ref{tab:efficiency-final} compares the two reward judges in the matched RL comparison: Qwen3-1.7B Probe and Qwen3-8B Generative.
The actor, training data, GRPO configuration, response budget, rubric aggregation rule, and external \gptfour{} evaluator are fixed; only the reward judge changes.
The RL reward Probe is trained once on 500 \gptfour{}-labeled responses (3,766 criterion labels; approximately \$2.43). The frozen Probe can then be reused for subsequent RL runs in the same rubric setting without further \gptfour{} calls or recalibration.
At reward time, the Probe requires substantially less judging computation.
By the comparison checkpoint, Generative requires 89,912.1 seconds of cumulative judge time versus 8,389.9 seconds for Probe, a 10.7$\times$ ratio.
On validation, Generative takes 492.0 seconds versus Probe's 31.1 seconds (15.8$\times$).

The efficiency result should be read with policy quality.
Probe reaches a higher \rar{} score, 0.643 versus 0.594, while using 2.33 judge-hours instead of 24.98 judge-hours.
The wall-clock reduction is smaller because reward calls are parallelized and actor rollout, reference-model scoring, optimization, and checkpointing remain in the loop.
The key result is not only lower judge cost: \textbf{the Probe reward produces the stronger policy while requiring substantially less reward-judge computation.}
Appendix~\ref{app:efficiency-measurement} gives the timing definitions and training-only calculation.

%% file: tables/matched_rl_results.tex
\begin{table}[t]
\centering
\footnotesize
\setlength{\tabcolsep}{3pt}
\resizebox{\columnwidth}{!}{%
\begin{tabular}{llcc}
\toprule
\textbf{Reward model} & \textbf{Method} & \textbf{\rar{} score} & \textbf{Score gain} \\
\midrule
Base actor, no RL & -- & 0.232 & -- \\
Qwen3-0.6B & Probe & 0.562 & +0.330 \\
Qwen3-1.7B & Probe & \textbf{0.643} & \textbf{+0.411} \\
Qwen3-4B & Probe & 0.588 & +0.356 \\
Qwen3-8B & Probe & 0.506 & +0.274 \\
Qwen3-8B & Generative & 0.594 & +0.362 \\
\bottomrule
\end{tabular}%
}
\caption{Matched Reward-model Comparison on \rar{}. All non-base rows train the same Qwen3-4B-Base actor with the same GRPO configuration and are evaluated by the same \gptfour{} rubric evaluator. The only experimental variable is the reward judge.}
\label{tab:rl-final}
\end{table}

%% file: tables/gpqa_transfer.tex
\begin{table}[t]
\centering
\footnotesize
\setlength{\tabcolsep}{5pt}
\begin{tabularx}{\columnwidth}{@{}Xcc@{}}
\toprule
\textbf{Policy} & \textbf{\gpqa{} acc.} & \textbf{$\Delta$} \\
\midrule
Base actor & 0.335 $\pm$ 0.023 & -- \\
Probe-reward policy & \textbf{0.388 $\pm$ 0.037} & +0.053 \\
\bottomrule
\end{tabularx}
\caption{Policy transfer to \gpqa{}. Accuracy is averaged over four evaluation runs on the 198-question \gpqa{} split, where in each run, the answer options are shown in a different order. The probe-reward policy is trained on \rar{} using the Qwen3-1.7B Probe reward.}

\label{tab:gpqa-transfer-final}
\end{table}

%% file: tables/cross_domain_transfer.tex
\begin{table}[t]
\centering
\footnotesize
\setlength{\tabcolsep}{4pt}
\begin{tabular}{llccc}
\toprule
\textbf{Train} & \textbf{Eval} & \textbf{Macro-F1} & \textbf{Score MAE} & \textbf{Pearson} \\
\midrule
Science & Science & 0.754 & 0.187 & 0.720 \\
Science & Medicine & 0.718 & \textbf{0.091} & \textbf{0.789} \\
Medicine & Medicine & \textbf{0.782} & 0.140 & 0.620 \\
Medicine & Science & 0.702 & 0.240 & 0.580 \\
\bottomrule
\end{tabular}
\caption{Static judge transfer across rubric domains. Probes are trained on source-domain criterion labels and evaluated against \gptfour{} criterion labels on the target domain without target-domain fitting.}
\label{tab:cross-domain-final}
\end{table}

%% file: tables/efficiency_quality.tex
\begin{table*}[!t]
\centering
\small
\setlength{\tabcolsep}{6.5pt}
\begin{tabular}{lrrrr}
\toprule
\textbf{Reward judge} & \textbf{Cumulative judge time} & \textbf{Validation judge time}
& \textbf{RL elapsed time} & \textbf{\rar{} score} \\
\midrule
Qwen3-1.7B Probe & 2.33h & 31.1s & 8h 51m 47s & \textbf{0.643} \\
Qwen3-8B Generative & 24.98h & 492.0s & 11h 11m 56s & 0.594 \\
\midrule
Generative / Probe ratio & 10.7$\times$ & 15.8$\times$ & 1.26$\times$ & -- \\
\bottomrule
\end{tabular}
\caption{Efficiency-quality comparison for the Qwen3-1.7B Probe judge and the generative baseline.
Judge time is measured through the same checkpoint used in Table~\ref{tab:rl-final}. Policy quality for both runs is measured by \gptfour{}. The Generative judge requires 10.7$\times$ as much cumulative judge time while producing a lower-scoring policy.}
\label{tab:efficiency-final}
\end{table*}

%% file: Sections/2-Related.tex
\section{Related Work}

\paragraph{Rubric-based RL.} Reinforcement learning with verifiable rewards works well when correctness can be checked by rule-based verifiers \citep{code-rl, grpo, rlvr, deepseek-r1}. Open-ended tasks require broader feedback, motivating rubric-style evaluation, rubric resources, and rubric-based RL \citep{healthbench, openrubrics, rar, dr-tulu}. \rar{} is closest to our downstream setting because it uses instance-specific science rubrics as scalar rewards for policy optimization \citep{rar}. Our work keeps the reward formulation fixed and studies the judge that applies the rubric, replacing repeated calls to a large judge with a smaller criterion-level evaluator.

\paragraph{LLM-as-a-judge and small evaluators.} LLM-as-a-judge methods use language models to score, compare, or critique open-ended responses \citep{llm-as-judge, liu-etal-2023-g, DBLP:conf/emnlp/MaZZLJQSCYV25, llm-judge-survey}. Trained open judges extend this idea by fine-tuning models for evaluation \citep{pandalm, auto-j, kim-etal-2024-prometheus, judgelm}. Examples of related rubric-oriented evaluators include LLM-Rubric, which studies calibrated multidimensional evaluation \citep{hashemi-etal-2024-llm}, and works such as FLAMe and Prometheus, which train open autoraters for rubric- or instruction-conditioned evaluation \citep{vu-etal-2024-foundational, kim2024prometheus}. Recent evaluator benchmarks study reward-model and judge reliability across evaluation formats \citep{lambert-etal-2025-rewardbench, DBLP:conf/nips/ZhaoZHWBLTCDBZH25, if-rewardbench}. Work on compact evaluators and hidden-state judges further suggests that smaller models can contain useful evaluative signals even when generation is weak \citep{DBLP:conf/emnlp/LiuF0LJWXR23, selene-mini, representation-judge, bert-as-a-judge}. Our work differs by keeping the language-model backbone frozen, training only a lightweight criterion classifier, and using its satisfaction probabilities as repeated rewards for rubric-based RL.

%% file: Sections/Appendix.tex
\clearpage
\appendix

\numberwithin{table}{section}

\section{\orb{} Construction and Rubric Scoring}
\label{app:additional-details}

\subsection{Pointwise and Pairwise Rubric Scoring}
\label{app:rubric-scoring}

Let $q$ denote a prompt, $y$ a candidate response, and $C(q)=\{c_j\}_{j=1}^{m_q}$ the rubric for $q$. Each criterion $c_j$ is associated with a nonnegative weight $w_j$, with $\sum_j w_j>0$ for every rubric. In settings with hard and soft criteria, $w_j$ is obtained by mapping the criterion type to a fixed weight; in settings with absolute rubric weights, $w_j$ is given directly by the rubric.
A pointwise rubric label is a criterion-satisfaction judgment
\[
    z_j(q,y) \in \{0,1\},
\]
where $z_j(q,y)=1$ means that response $y$ satisfies criterion $c_j$ for prompt $q$. A probabilistic judge instead returns $p_j(q,y)\in[0,1]$, interpreted as the probability that $c_j$ is satisfied.

The normalized rubric score for a response is
\[
    R(q,y) =
    \frac{\sum_{j=1}^{m_q} w_j z_j(q,y)}
         {\sum_{j=1}^{m_q} w_j},
\]
or, for probabilistic judges,
\[
    \widehat{R}(q,y) =
    \frac{\sum_{j=1}^{m_q} w_j p_j(q,y)}
         {\sum_{j=1}^{m_q} w_j}.
\]
This aggregation preserves the rubric structure while producing the scalar reward required for reinforcement learning.

A pairwise preference between two responses $y_a$ and $y_b$ to the same prompt can be induced from pointwise rubric scores:
\[
    y_a \succ_q y_b
    \quad \Longleftrightarrow \quad
    R(q,y_a) > R(q,y_b),
\]
with ties defined when the two normalized scores are equal. Pairwise labels are therefore recoverable from pointwise rubric labels when the same rubric is applied to both responses. The reverse is not true: a pairwise preference does not identify which criteria were satisfied or violated. Pointwise labels are therefore richer for rubric-based RL because they preserve the criterion-level reward structure used to compute rewards, analyze criterion-level errors, and train reward readouts without collapsing the rubric to a single preference label.

For example, suppose a rubric contains one criterion requiring the final answer and another requiring supporting reasoning. A response may satisfy the final answer criterion while failing the reasoning criterion. Pointwise labels retain this distinction, whereas a pairwise label only states whether that response is preferred to another response under the aggregate score.

\subsection{\orb{} Construction Details}
\label{app:orb-construction-details}

Table~\ref{tab:orb-pipeline} summarizes the construction procedure. We sample question records from \openrubrics{} \citep{openrubrics} whose original rubrics contain explicit hard requirements, then refine each rubric into a fixed six-criterion format with two hard criteria and four soft criteria. We use \gptfour{} for rubric refinement and reference grading. The final benchmark is obtained by retaining questions whose generated response set spans full-score, low-score, and intermediate rubric outcomes.

\begin{table}[t]
\centering
\footnotesize
\setlength{\tabcolsep}{3pt}
\begin{tabularx}{\columnwidth}{@{}l r X@{}}
\toprule
Stage & Count & Criterion \\
\midrule
Candidates & 3,000 & Two sampled construction passes \\
Unique questions & 2,902 & Before filtering \\
Full-score answer & 2,432 & Answer 1 score $=10$ \\
Low-score answer & 1,566 & Also answer 2 score $\leq 3$ \\
Intermediate answer & 1,061 & Also some answer score in $[2,8]$ \\
Unique after filters & 1,045 & Deduplicated by question ID \\
Frozen \orb{} & 1,042 & Canonical benchmark used in all experiments \\
\bottomrule
\end{tabularx}
\caption{\orb{} construction stages. Counts are computed over completed construction artifacts. The final row gives the frozen benchmark artifact used for all experiments.}
\label{tab:orb-pipeline}
\end{table}

\subsection{Rubric-Transformation Audit}
\label{app:orb-transformation-audit}

We manually audited 50 transformations sampled uniformly from the construction data with seed 42. For each example, we compared the original question and \openrubrics{} Hard Rules and Principles with the standardized two-hard/four-soft rubric, without reference to generated responses or criterion labels. The audit assessed whether the transformation preserved the core rubric intent, avoided unsupported requirements, and produced a reasonable hard/soft decomposition.
Because the transformation compresses rubrics into a fixed structure, preservation was evaluated at the level of substantive intent rather than one-to-one retention of every original Principle.

Of the 50 transformations, 47 preserved the core rubric intent, all 50 avoided unsupported additions, and 49 produced a reasonable hard/soft decomposition.
The three intent-preservation failures involved requirements omitted during compression rather than newly introduced requirements.

\paragraph{Representative transformations.}
In a faithful Python/Celery example, the original hard rules require activating a Python 3.8 environment and providing the exact Celery 5.2.7 installation command with Redis support. The standardized rubric retains both requirements as hard criteria and consolidates the original compatibility, troubleshooting, and verification Principles into four soft criteria without changing the core grading intent.

In an imperfect browser-game example, the original rubric explicitly requires the core interactive mechanics of a Rocket League--style implementation. The standardized rubric retains the browser implementation, dependency, structure, usability, and code-correctness requirements, but does not retain an explicit criterion for those mechanics. We therefore mark this transformation as not preserving the complete rubric intent. These examples illustrate both the intended compression and its principal failure mode.

\subsection{\orb{} Response Roles}
\label{app:orb-response-roles}

Table~\ref{tab:orb-response-roles} describes the four responses associated with each question. The responses should not be interpreted as four ordinal quality levels. Instead, \orb{} fixes one full-score response and one low-score response, then includes two additional comparison responses whose observed scores vary.

\begin{table*}[t]
\centering
\small
\begin{tabularx}{\textwidth}{c l l X X}
\toprule
ID & Role & Rubric visible & Generator in final artifact & Retention status \\
\midrule
1 & Full-score response & Yes & \gptfour{} (1,042) &
Required to receive total score 10. \\
2 & Low-score response & Yes & \gptfour{} (1,042) &
Required to receive total score at most 3. \\
3 & Partial-compliance response & Yes & \gptfour{} (764), GPT-4.1-nano \citep{gpt41} (278) &
Additional comparison response; broad observed score range. \\
4 & Unguided response & No & GPT-4.1-nano (1,042) &
Additional comparison response; natural score variation without rubric access. \\
\bottomrule
\end{tabularx}
\caption{Response roles in \orb{}. Counts reflect the final benchmark artifact.}
\label{tab:orb-response-roles}
\end{table*}

\begin{table}[t]
\centering
\small
\begin{tabular}{c r r r r r}
\toprule
ID & Mean & 0--3 & 4--6 & 7--9 & 10 \\
\midrule
1 & 10.00 & 0 & 0 & 0 & 1,042 \\
2 & 0.89 & 1,042 & 0 & 0 & 0 \\
3 & 6.39 & 89 & 461 & 367 & 125 \\
4 & 8.47 & 44 & 117 & 395 & 486 \\
\bottomrule
\end{tabular}
\caption{Observed \gptfour{} rubric-score distribution by response ID in the final \orb{} benchmark. Scores range from 0 to 10 under hard-rule weight 3 and soft-rule weight 1.}
\label{tab:orb-score-distribution}
\end{table}

\subsection{\orb{} Evaluation Protocol and Metrics}
\label{app:orb-protocol-metrics}

For judge methods that require training or calibration, we use a fixed question-level split so that responses to the same question never appear in both calibration and held-out evaluation. Frozen or zero-shot judges can be evaluated on the full benchmark.

\begin{table}[t]
\centering
\small
\begin{tabular}{@{}lrrr@{}}
\toprule
Split & Questions & Responses & Criterion labels \\
\midrule
Train & 417 & 1,668 & 10,008 \\
Development & 104 & 416 & 2,496 \\
Held-out & 521 & 2,084 & 12,504 \\
\midrule
Total & 1,042 & 4,168 & 25,008 \\
\bottomrule
\end{tabular}
\caption{Question-level \orb{} split used for calibrated or trained judge methods, after validating all criterion labels.}
\label{tab:orb-split}
\end{table}

The primary \orb{} metric is weighted criterion accuracy, using weight 3 for hard criteria and weight 1 for soft criteria. Complementary static metrics include parse success, unweighted criterion accuracy, macro-F1, balanced accuracy, MCC, scalar-score error, score correlation, and induced pairwise ordering accuracy.

\subsection{Reference Judge and Construction Prompts}
\label{app:orb-reference-judge}

We use \gptfour{} \citep{gpt4o} as the construction and reference judge for \orb{}. During benchmark construction, GPT-5 \citep{gpt5} frequently refused or failed to produce responses with the designated quality level, especially deliberately low-scoring or partially compliant answers. Since \orb{} requires controlled variation across responses to the same prompt, we used \gptfour{} for rubric refinement, designated-quality response generation, and criterion-level grading. For consistency, all subsequent \orb{} judge comparisons use the same \gptfour{} reference labels. The rubric-refinement, answer-generation, and grading prompts are provided for reproducibility in Appendix~\ref{app:prompts}.

\subsection{\rar{}-Static Response Bank and Splits}
\label{app:rar-static-bank}

\rar{}-Static is the fixed response bank used for static judge evaluation under the \rar{} rubric format. It contains 1,500 science questions sampled from \rar{} \citep{rar}. Each question is paired with two candidate responses: the original dataset reference answer and a greedy Qwen3-4B response \citep{qwen3}. The resulting bank contains 3,000 candidate responses. The rubrics are instance-specific; in \rar{}-Static, they contain between 6 and 11 criteria, with an average of 7.52 criteria per question.

For trained or calibrated static methods, we use a fixed question-level split: 1,000 training questions, 200 development questions, and 300 held-out questions.
The held-out split contains 600 candidate responses. Splitting is performed by question, so no responses to the same question appear in both calibration and evaluation. Generative and Logprob judges do not require fitting, but they are evaluated on the same held-out questions for comparability.

\paragraph{Extended held-out response bank.}
\label{app:rar-extended-bank}
To test response-generator shift, we extend the 300-question held-out bank with greedy responses from Mistral-7B-Instruct-v0.3 and OLMo-2-1124-7B-Instruct \citep{mistral7b,olmo2}. Together with the original reference answer and greedy Qwen3-4B response, this gives four responses per question and 1,200 response-level examples. All added responses are labeled by \gptfour{} using the same criterion-level protocol. The training and development data, Probe parameters, layer, and threshold remain unchanged.

\begin{table}[t]
\centering
\footnotesize
\setlength{\tabcolsep}{3pt}
\resizebox{\columnwidth}{!}{%
\begin{tabular}{@{}lrrrrr@{}}
\toprule
Judge & Macro-F1 & Agreement & Pearson & MAE & Parse rate \\
\midrule
Generative & 0.394 & 0.523 & 0.130 & 0.471 & 0.950 \\
Logprob & 0.424 & 0.538 & 0.404 & 0.405 & 1.000 \\
Probe & \textbf{0.741} & \textbf{0.746} & \textbf{0.683} & \textbf{0.224} & 1.000 \\
\bottomrule
\end{tabular}%
}
\caption{Qwen3-1.7B judge results on the extended \rar{}-Static held-out bank.
Agreement is criterion-level agreement; Pearson and MAE are computed over weighted scalar rubric scores.}
\label{tab:rar-extended-bank}
\end{table}

The extended bank is more difficult than the original two-response bank, but the readout ranking remains unchanged. The unchanged Probe retains stronger criterion agreement, scalar-score correlation, and score accuracy on responses from the additional model families.

\subsection{\rar{}-Static Reference Labels and Aggregation}
\label{app:reference-aggregation}

We use \gptfour{} \citep{gpt4o} as the operational reference judge for \rar{}-Static. The reference judge assigns one Yes/No verdict per rubric criterion for each candidate response. The final reference-scoring run parses successfully for all 3,000 candidate responses.

For scalar-score metrics, \rar{}-Static uses the original \rar{} rubric weights after converting all criteria into a satisfaction-label interface. Some \rar{} criteria are pitfall criteria with negative raw weights. We interpret these as avoidance criteria: satisfying the criterion means that the response avoids the pitfall. We therefore aggregate with absolute weights so that higher criterion satisfaction always corresponds to better rubric compliance:
\[
    \widehat{R}(q,y)=
    \frac{\sum_{j=1}^{m_q} |w_j|\,p_j(q,y)}
         {\sum_{j=1}^{m_q} |w_j|}.
\]
For binary methods, $p_j(q,y)$ is replaced by $\hat{z}_j(q,y)$. The same aggregation rule is used for Generative, Logprob, and Probe judges on \rar{}-Static.

\subsection{Reference-Label Validation}
\label{app:dataset-validation}

We validate the \gptfour{} reference labels with a blinded human audit on the held-out \orb{} split used in our static evaluator experiments. The audit samples 50 question-level groups, and for each group selects two candidate responses from the four available responses such that the two candidates have different \gptfour{} rubric scores. The resulting 100 responses are shuffled before annotation. For each response, the annotator labels every rubric criterion as satisfied or not satisfied without seeing the \gptfour{} labels or scores.

The audit covers 600 response--criterion decisions. We compute criterion-level agreement between the human annotator and \gptfour{} labels, and additionally report a weighted agreement that uses the rubric criterion weights when aggregating decisions. Human labels agree with \gptfour{} on 90.9\% of weighted criterion decisions. Agreement is highest for hard factual or constraint-following criteria, and lower for softer criteria involving completeness, clarity, specificity, and overall explanation quality. Table~\ref{tab:human-audit} summarizes the audit size and agreement results.

We also evaluate whether the human audit preserves the response ordering induced by the rubric score. For each of the 50 sampled response pairs, we aggregate the human criterion labels with the same rubric weights and compare the preferred response with the \gptfour{}-preferred response. Human and \gptfour{} preferences agree on 96.0\% of pairs, further supporting the reliability of the operational reference target.

\begin{table}[t]
\centering
\small
\setlength{\tabcolsep}{8pt}
\begin{tabular}{lr}
\toprule
Audit statistic & Value \\
\midrule
Sampled responses & 100 \\
Sampled response pairs & 50 \\
Response--criterion decisions & 600 \\
Weighted criterion agreement with \gptfour{} & 90.9\% \\
Pairwise preference agreement with \gptfour{} & 96.0\% \\
\bottomrule
\end{tabular}
\caption{Human audit of \gptfour{} reference labels on the held-out \orb{} split.}
\label{tab:human-audit}
\end{table}

\section{Static Evaluation Details}

\subsection{Generative and Logprob Judge Details}
\label{app:generated-logprob-details}

\paragraph{Generative judging.}
The Generative judge receives one criterion at a time and produces a single Yes/No verdict. Verdicts are then recombined into a full criterion vector for the candidate response. This reduces the output-format burden for each completion but requires one generation per criterion.

\paragraph{Forced-choice Logprob scoring.}
The Logprob judge receives the question, candidate response, and one canonicalized criterion. The prompt asks whether the response satisfies the criterion and ends before the answer token. We score the next-token probabilities of the fixed verbalizers `` Yes'' and `` No'' and compute their log-probability margin:
\[
    \Delta_j=\log P(\text{ Yes}\mid q,y,c_j)-\log P(\text{ No}\mid q,y,c_j).
\]
We set the soft satisfaction score to \(p_j(q,y)=\sigma(\Delta_j)\). Binary metrics threshold the margin at 0, equivalently \(p_j(q,y)\geq 0.5\). Scalar rubric aggregation uses the soft satisfaction score.

\subsection{Probe Training and Calibration}
\label{app:probe-details}

\paragraph{Canonical criterion rendering.}
Probe prompts use a single criterion at a time. Ordinary criteria are rendered as satisfaction criteria. Pitfall criteria are rendered as avoidance criteria, so a positive label always means that the response complies with the rubric.

\paragraph{Representation extraction.}
For each rendered prompt, we run the frozen judge model with hidden states enabled and extract the representation at the final non-padding token. For layer $\ell$, this gives $h_\ell(q,y,c_j)$. We evaluate last-token, mean-pooled, and last-few-token pooled representations in the ablation table.

\paragraph{Feature standardization.}
For each layer, training features are used to compute a mean vector and standard deviation vector. Training, development, and held-out features are standardized with these training-set statistics. Held-out features are never used to choose normalization statistics, layers, thresholds, or classifier checkpoints.

\paragraph{Probe heads.}
We train linear and MLP binary classifiers with binary cross-entropy. The classifier predicts the \gptfour{} criterion label from the frozen representation. For MLP probes, only the small probe head is trained; the language model remains frozen.

\paragraph{Layer and threshold selection.}
For each candidate layer, we train the probe on the training split and evaluate on the development split. The threshold is selected on development data to maximize macro-F1, and the layer is selected by the same development metric. The held-out split is evaluated only after these choices are fixed.

\begin{table}[t]
\centering
\footnotesize
\setlength{\tabcolsep}{4pt}
\begin{tabularx}{\columnwidth}{@{}lX@{}}
\toprule
Training detail & Setting \\
\midrule
Backbone & Frozen language model \\
Representation & Final non-padding-token hidden state \\
Head & Linear; MLP only in ablations \\
Optimization & Full-batch AdamW, 200 epochs; learning rate 0.01; weight decay 0.01 \\
Objective & Binary cross-entropy with positive-class weighting \\
Selection & Head checkpoint by development loss; layer and threshold by development macro-F1 \\
Seed & Split seed 42; head seed 42 plus layer index \\
\bottomrule
\end{tabularx}
\caption{Probe fitting and selection settings.}
\label{tab:probe-training-settings}
\end{table}

\paragraph{Supervision cost and reuse.}
Static evaluation and RL use separate supervision banks. The main \rar{}-Static Probe uses 2,400 responses (18,032 criterion labels; approximately \$11.62), whereas the RL reward Probe uses 450 training and 50 development responses (3,766 labels; approximately \$2.43). Once fitted, either frozen Probe can be reused in the same rubric setting without further teacher calls or recalibration. The complete \gptfour{} reference-labeling run used 4.11 million tokens and cost \$14.53 at the API pricing used during construction.

The data-size ablation indicates that smaller supervision sets remain viable: 500 training questions achieve 0.828 macro-F1, compared with 0.834 for 1,000 questions, while 100 questions achieve 0.805. The science-to-medicine transfer result suggests useful agreement under a moderate rubric shift without refitting, but substantial changes in domain or rubric format may require new supervision. Across the studied training sizes, the corresponding reference-labeling cost ranges from approximately \$2.91 to \$11.62. These estimates use the API pricing at the time of dataset construction.

\paragraph{Probability rewards.}
At evaluation time, the probe outputs a criterion probability. Scalar scores use these probabilities rather than only thresholded verdicts unless otherwise specified. In the RL experiments, these probabilities provide the reward signal while preserving the same explicit rubric aggregation used by generative judges.

\subsection{Supervised Fine-tuning Details}
\label{app:sft-details}

SFT uses the same train, development, and held-out question splits as the corresponding static evaluation setting. The model is fine-tuned only on the training split, with development loss used for checkpoint selection. Held-out questions are used only for final reporting. After SFT, the resulting backbone is evaluated with the same Generative, Logprob, and Probe methods used for the base backbone.

\begin{table}[t]
\centering
\footnotesize
\setlength{\tabcolsep}{4pt}
\begin{tabularx}{\columnwidth}{@{}XX@{}}
\toprule
Metric & Value \\
\midrule
Backbones & Qwen3-0.6B, Qwen3-1.7B, Qwen3-4B, Qwen3-8B \citep{qwen3} \\
Training objective & Supervised verdict prediction \\
Fine-tuning method & LoRA \citep{lora} \\
LoRA target modules & All target modules \\
Chat template & Qwen \\
Precision & bf16 \\
Optimizer & AdamW \citep{adamW} \\
Scheduler & Cosine decay \\
Warmup ratio & 0.03 \\
Epochs & 3 \\
Checkpoint selection & Lowest development loss \\
Checkpoint limit & 3 \\
\midrule
\orb{} training examples & Question-level training split \\
\orb{} maximum length & 4096 tokens \\
\orb{} learning rate & $2\times10^{-4}$ \\
\orb{} eval/save frequency & 200 steps \\
\orb{} batch/accumulation & 0.6B: batch 2, accumulation 16; larger models: batch 1, accumulation 32 \\
\midrule
\rar{}-Static training examples & Question-level training split \\
\rar{}-Static maximum length & 8192 tokens \\
\rar{}-Static learning rate & $1\times10^{-4}$ \\
\rar{}-Static eval/save frequency & 50 steps \\
\rar{}-Static batch/accumulation & 0.6B: batch 2, accumulation 16; larger models: batch 1, accumulation 16 \\
\bottomrule
\end{tabularx}
\caption{SFT configuration for the post-training judge experiments.}
\label{tab:sft-config}
\end{table}

In both data settings, SFT trains the model to output the complete rubric verdict vector for a candidate response. Each example contains the question, candidate response, all rubric criteria, and the corresponding \gptfour{} Yes/No verdicts.
\rar{}-Static examples use responses and verdicts from the fixed response bank.
For SFT Probe, the Probe is trained on hidden states extracted from the SFT backbone using the same Probe-training protocol as for the base backbone.

\subsection{Static Evaluation Targets}
\label{app:static-targets}

For both \orb{} and \rar{}, the static target is criterion satisfaction. Each example is a tuple $(q,y,c_j)$, where $q$ is the question, $y$ is a candidate response, and $c_j$ is one rubric criterion. The operational reference label is
\[
    z_j^*(q,y) \in \{0,1\},
\]
where $z_j^*(q,y)=1$ means that \gptfour{} \citep{gpt4o} judges response $y$ to satisfy criterion $c_j$ for question $q$.

A judge method may return either a binary verdict $\hat{z}_j(q,y)\in\{0,1\}$ or a probability $p_j(q,y)\in[0,1]$. Binary metrics are computed from $\hat{z}_j(q,y)$, using a thresholded probability when the method is probabilistic. Scalar rubric scores are computed by aggregating criterion predictions with the rubric weights, so the scalar score is derived from criterion-level judgments rather than predicted directly.

\subsection{Rubric-RM Pairwise Comparison}
\label{app:rubric-rm-pairwise}

Rubric-RM directly predicts the preferred response from a pair conditioned on the rubric \citep{openrubrics}. By contrast, our pointwise judges score each response independently, after which their scalar scores induce a pairwise ordering. We compare the methods on all pairs for which the \gptfour{} reference scores are non-tied. Because Rubric-RM sometimes fails to produce a parseable preference under this adaptation, we report its parse rate, accuracy over the complete evaluation set with parse failures counted as incorrect, and accuracy restricted to parseable outputs.

\begin{table}[t]
\centering
\footnotesize
\setlength{\tabcolsep}{3pt}
\resizebox{\columnwidth}{!}{%
\begin{tabular}{@{}llrrr@{}}
\toprule
Dataset & Method & Parse & Full set & Parseable \\
\midrule
\orb{} & Rubric-RM-4B & 0.627 & 0.322 & 0.513 \\
\orb{} & Rubric-RM-8B & 0.396 & 0.228 & 0.575 \\
\orb{} & Qwen3-1.7B Probe & 1.000 & \textbf{0.924} & \textbf{0.924} \\
\rar{}-Static & Rubric-RM-4B & 0.727 & 0.487 & 0.670 \\
\rar{}-Static & Rubric-RM-8B & 0.401 & 0.210 & 0.523 \\
\rar{}-Static & Qwen3-1.7B Probe & 1.000 & \textbf{0.888} & \textbf{0.888} \\
\bottomrule
\end{tabular}%
}
\caption{Static pairwise comparison on \gptfour{} non-tied pairs. Full-set accuracy counts parse failures as incorrect; parseable accuracy excludes them.}
\label{tab:rubric-rm-pairwise}
\end{table}

For \orb{}, the Probe produces 287 exact score ties; its reported accuracy assigns half credit to these ties. The \rar{}-Static Probe produces no ties.
Rubric-RM's parseable-only results show that it captures useful pairwise signal when its output conforms to the requested format, but parse failures substantially reduce full-set accuracy.

This comparison is not a direct online-RL efficiency comparison. With a rollout group of size $k$, a pairwise reward model requires up to $k(k-1)/2$ comparisons, whereas the pointwise interface requires $k$ response scores and additionally exposes criterion-level satisfaction. At our group size of four, this corresponds to six pairwise comparisons rather than four pointwise response evaluations.

\subsection{Static Readout Error Analysis}
\label{app:static-readout-errors}

We compare the Qwen3-1.7B Generative and Probe predictions on the same 4,518 \rar{}-Static held-out criterion decisions. The paired decomposition separates errors recovered by the Probe, errors introduced by the Probe, and criteria that remain difficult for both readouts.

\begin{table}[t]
\centering
\footnotesize
\setlength{\tabcolsep}{3pt}
\begin{tabularx}{\columnwidth}{@{}Xlr@{}}
\toprule
Analysis & Outcome & Share \\
\midrule
Paired & Generative wrong, Probe correct & 33.0\% \\
Paired & Generative correct, Probe wrong & 7.8\% \\
Paired & Both wrong & 7.9\% \\
Paired & Both correct & 51.3\% \\
\midrule
Error & Generative false positive & 33.7\% \\
Error & Generative false negative & 4.2\% \\
Error & Probe false positive & 6.7\% \\
Error & Probe false negative & 9.0\% \\
\bottomrule
\end{tabularx}
\caption{Paired Qwen3-1.7B readout errors on the \rar{}-Static held-out split.
Shares are computed over 4,518 criterion decisions.}
\label{tab:static-readout-errors}
\end{table}

The Generative readout's errors are dominated by false positives, indicating a tendency to accept criteria that the \gptfour{} reference marks as unsatisfied.
The Probe removes many of these false positives, although its remaining errors are more balanced between false positives and false negatives. The 7.8\% reverse outcome also shows that the Probe does not uniformly dominate generation at the individual-decision level.

\paragraph{Representative Generative false positive.}
For the question asking for the principal energy-liberating process in stars, the candidate answers only, ``Conversion of hydrogen into helium via nuclear fusion.'' One criterion additionally requires explaining that fusion releases energy by converting mass into energy. \gptfour{} and Probe mark this criterion unsatisfied, while Generative marks it satisfied. This example illustrates the Generative readout's tendency to accept criteria whose central topic is mentioned but whose specific requirement is omitted.

\paragraph{Representative Probe false negative.}
For the bound-state problem with $V(x)=-a\delta(x)$, the candidate explicitly gives a wavefunction that decays exponentially on both sides of $x=0$.
\gptfour{} and Generative therefore mark the wavefunction-form criterion satisfied, while Probe marks it unsatisfied. This case shows that Probe can miss requirements that are directly supported by the response.

\subsection{Static Metrics}
\label{app:static-metrics}

\begin{table*}[!t]
\centering
\scriptsize
\setlength{\tabcolsep}{4pt}
\resizebox{\textwidth}{!}{%
\begin{tabular}{rrrrrr}
\toprule
Layer & Dev Macro-F1 & Held-out Acc. & Held-out Macro-F1 & Held-out Bal. Acc. & Held-out MCC \\
\midrule
1 & 0.726 & 0.759 & 0.738 & 0.733 & 0.480 \\
2 & 0.742 & 0.774 & 0.750 & 0.743 & 0.508 \\
3 & 0.751 & 0.781 & 0.754 & 0.745 & 0.523 \\
4 & 0.752 & 0.787 & 0.769 & 0.764 & 0.540 \\
5 & 0.755 & 0.778 & 0.751 & 0.742 & 0.517 \\
6 & 0.752 & 0.791 & 0.771 & 0.763 & 0.548 \\
7 & 0.758 & 0.791 & 0.779 & 0.780 & 0.558 \\
8 & 0.768 & 0.794 & 0.774 & 0.766 & 0.554 \\
9 & 0.765 & 0.787 & 0.774 & 0.775 & 0.548 \\
10 & 0.778 & 0.804 & 0.790 & 0.788 & 0.581 \\
11 & 0.784 & 0.810 & 0.800 & 0.802 & 0.600 \\
12 & 0.790 & 0.819 & 0.805 & 0.800 & 0.611 \\
13 & 0.805 & 0.827 & 0.812 & 0.805 & 0.627 \\
14 & 0.799 & 0.830 & 0.818 & 0.814 & 0.636 \\
15 & 0.814 & 0.831 & 0.814 & 0.806 & 0.635 \\
\textbf{16} & \textbf{0.836} & \textbf{0.845} & \textbf{0.833} & \textbf{0.829} & \textbf{0.667} \\
17 & 0.836 & 0.849 & 0.839 & 0.837 & 0.678 \\
18 & 0.832 & 0.853 & 0.841 & 0.835 & 0.684 \\
19 & 0.827 & 0.847 & 0.835 & 0.829 & 0.672 \\
20 & 0.824 & 0.843 & 0.835 & 0.838 & 0.670 \\
21 & 0.828 & 0.845 & 0.835 & 0.836 & 0.671 \\
22 & 0.827 & 0.843 & 0.834 & 0.833 & 0.667 \\
23 & 0.823 & 0.848 & 0.837 & 0.835 & 0.675 \\
24 & 0.821 & 0.847 & 0.837 & 0.837 & 0.675 \\
25 & 0.819 & 0.843 & 0.834 & 0.835 & 0.668 \\
26 & 0.818 & 0.844 & 0.835 & 0.835 & 0.670 \\
27 & 0.816 & 0.839 & 0.828 & 0.825 & 0.656 \\
28 & 0.817 & 0.843 & 0.833 & 0.832 & 0.665 \\
\bottomrule
\end{tabular}
}
\caption{Qwen3-1.7B Probe layer sweep on \rar{}-Static. Layer selection is based on development macro-F1; the selected layer is bolded.}
\label{tab:probe-layer-sweep-full}
\end{table*}

\begin{table*}[t]
\centering
\scriptsize
\setlength{\tabcolsep}{3pt}
\resizebox{\textwidth}{!}{%
\begin{tabular}{lcccccc}
\toprule
& \multicolumn{3}{c}{Base model} & \multicolumn{3}{c}{SFT model} \\
\cmidrule(lr){2-4}\cmidrule(lr){5-7}
Model & Generative & Logprob & Probe & Generative & Logprob & Probe \\
\midrule
Qwen3-0.6B & 0.370 [0.349, 0.390] & 0.582 [0.568, 0.596] & 0.802 [0.791, 0.811] & 0.560 [0.537, 0.584] & 0.604 [0.591, 0.618] & 0.820 [0.810, 0.830] \\
Qwen3-1.7B & 0.518 [0.493, 0.541] & 0.766 [0.757, 0.776] & 0.875 [0.867, 0.883] & 0.902 [0.894, 0.909] & 0.713 [0.705, 0.722] & 0.845 [0.836, 0.854] \\
Qwen3-4B & 0.790 [0.768, 0.811] & 0.893 [0.886, 0.899] & 0.913 [0.906, 0.919] & 0.808 [0.788, 0.827] & 0.892 [0.885, 0.899] & 0.906 [0.899, 0.912] \\
Qwen3-8B & 0.888 [0.876, 0.899] & 0.907 [0.900, 0.914] & 0.914 [0.907, 0.921] & 0.846 [0.828, 0.862] & 0.895 [0.889, 0.902] & 0.912 [0.906, 0.918] \\
\bottomrule
\end{tabular}
}
\caption{\orb{} static results with 95\% bootstrap confidence intervals. Values are weighted criterion accuracy.}
\label{tab:orb-static-ci}
\end{table*}

\begin{table*}[t]
\centering
\scriptsize
\setlength{\tabcolsep}{3pt}
\resizebox{\textwidth}{!}{%
\begin{tabular}{lcccccc}
\toprule
& \multicolumn{3}{c}{Base model} & \multicolumn{3}{c}{SFT model} \\
\cmidrule(lr){2-4}\cmidrule(lr){5-7}
Model & Generative & Logprob & Probe & Generative & Logprob & Probe \\
\midrule
Qwen3-0.6B & 0.413 [0.398, 0.427] & 0.433 [0.414, 0.451] & 0.793 [0.775, 0.811] & 0.607 [0.512, 0.695] & 0.605 [0.583, 0.626] & 0.796 [0.778, 0.813] \\
Qwen3-1.7B & 0.443 [0.426, 0.459] & 0.449 [0.431, 0.467] & 0.835 [0.819, 0.851] & 0.708 [0.661, 0.753] & 0.520 [0.497, 0.544] & 0.828 [0.813, 0.843] \\
Qwen3-4B & 0.496 [0.475, 0.518] & 0.738 [0.716, 0.760] & 0.851 [0.836, 0.865] & 0.673 [0.637, 0.707] & 0.794 [0.776, 0.812] & 0.845 [0.829, 0.859] \\
Qwen3-8B & 0.609 [0.587, 0.632] & 0.756 [0.736, 0.775] & 0.864 [0.850, 0.878] & 0.642 [0.598, 0.684] & 0.754 [0.735, 0.774] & 0.861 [0.847, 0.874] \\
\bottomrule
\end{tabular}
}
\caption{\rar{}-Static results with 95\% bootstrap confidence intervals. Values are criterion-level macro-F1.}
\label{tab:rar-static-ci}
\end{table*}

\begin{table}[t]
\centering
\footnotesize
\setlength{\tabcolsep}{2pt}
\begin{tabularx}{\columnwidth}{@{}llc@{}}
\toprule
Group & Setting & Macro-F1 [95\% CI] \\
\midrule
Data size & 50 & 0.767 [0.747, 0.785] \\
Data size & 100 & 0.805 [0.787, 0.820] \\
Data size & 250 & 0.818 [0.801, 0.834] \\
Data size & 500 & 0.828 [0.811, 0.843] \\
Data size & 1000 & 0.834 [0.818, 0.849] \\
\midrule
Probe design & Linear, last token & 0.834 [0.818, 0.849] \\
Probe design & MLP-32, last token & 0.834 [0.818, 0.850] \\
Probe design & MLP-64, last token & 0.840 [0.824, 0.855] \\
Probe design & Linear, mean pooled & 0.761 [0.739, 0.782] \\
Probe design & Linear, last-4-layer mean & 0.839 [0.823, 0.854] \\
\bottomrule
\end{tabularx}
\caption{Qwen3-1.7B probe ablations on \rar{}-Static with 95\% bootstrap confidence intervals.}
\label{tab:probe-ablation-ci}
\end{table}

\paragraph{Parse success.}
For generative judges, sample-level parse success is the fraction of candidate responses for which the judge output can be parsed into a complete criterion verdict vector. Criterion-level parse success is the fraction of individual criterion verdicts successfully recovered. Logprob and probe methods are defined without generative verdict parsing and therefore produce scores for every rendered criterion.

\paragraph{Criterion accuracy.}
Criterion accuracy is the Hamming accuracy over criterion labels:
\[
    \mathrm{Acc}
    =
    \frac{1}{N}
    \sum_{(q,y,j)}
    \mathbbm{1}\{\hat{z}_j(q,y)=z_j^*(q,y)\}.
\]

\paragraph{Weighted criterion accuracy.}
Weighted criterion accuracy weights each criterion match by the rubric weight:
\[
    \mathrm{WAcc}
    =
    \frac{
    \sum_{(q,y,j)} \omega_{qj}
    \mathbbm{1}\{\hat{z}_j(q,y)=z_j^*(q,y)\}
    }{
    \sum_{(q,y,j)} \omega_{qj}
    }.
\]
For \orb{}, $\omega_{qj}$ follows the fixed hard/soft weights. For \rar{}-Static, $\omega_{qj}=|w_{qj}|$, since criteria and weights are question-specific.

\paragraph{Macro-F1.}
Macro-F1 is computed over the two criterion labels, Yes and No. Let $F_1^{\mathrm{Yes}}$ be the F1 score treating satisfied criteria as the positive class, and let $F_1^{\mathrm{No}}$ be the F1 score treating unsatisfied criteria as the positive class. Then
\[
    \mathrm{MacroF1}
    =
    \frac{1}{2}
    \left(
    F_1^{\mathrm{Yes}} + F_1^{\mathrm{No}}
    \right).
\]
We use macro-F1 as the headline \rar{} static metric because the distribution of satisfied and unsatisfied criteria varies across questions and candidate types.

\paragraph{Balanced accuracy and MCC.}
Balanced accuracy averages the recall of the Yes and No classes, making it less sensitive to class imbalance than ordinary accuracy. Matthews correlation coefficient (MCC) is computed from the binary confusion matrix and summarizes criterion-level agreement on a scale from $-1$ to $1$.

\paragraph{Scalar-score error.}
For each candidate response, we compare the judge's aggregated score $\widehat{R}(q,y)$ with the \gptfour{} aggregated score $R^*(q,y)$. Mean absolute error is
\[
    \mathrm{MAE}
    =
    \frac{1}{M}
    \sum_{(q,y)}
    |\widehat{R}(q,y)-R^*(q,y)|.
\]
We also report RMSE when useful.

\paragraph{Score correlation.}
Pearson correlation measures linear agreement between predicted and reference scalar scores:
\[
    r =
    \frac{\sum_i(\widehat{R}_i-\bar{\widehat{R}})
                  (R_i^*-\bar{R}^*)}
         {\sqrt{\sum_i(\widehat{R}_i-\bar{\widehat{R}})^2}
          \sqrt{\sum_i(R_i^*-\bar{R}^*)^2}}.
\]
Spearman correlation is the same correlation computed over average ranks rather than raw scores.

\paragraph{Pairwise ordering.}
When two candidate responses are available for the same question, we compare the ordering induced by the judge's scalar scores with the ordering induced by \gptfour{} scores. A pair is correct if both judges prefer the same response, or if both assign a tie.

\subsection{Bootstrap Confidence Intervals}
\label{app:bootstrap-ci}

We compute 95\% bootstrap confidence intervals by resampling questions with replacement. For \orb{}, each resampled question contributes its four candidate responses; for \rar{}, each resampled question contributes its two candidate responses. Tables~\ref{tab:orb-static-ci} and \ref{tab:rar-static-ci} report intervals for the main static judge results, and Table~\ref{tab:probe-ablation-ci} reports intervals for the probe ablations.

\subsection{Probe Layer Selection}
\label{app:probe-layer-selection}

The follow-up all-layer sweep selects layer 16 and obtains 0.833 held-out macro-F1, closely matching 0.834--0.835 for the other Probe fits. The similar performance of nearby middle-to-late layers indicates a broad plateau rather than dependence on one favorable layer.

\section{Rubric RL Details}
\label{app:additional-results-tables}

\begin{table}[t]
\centering
\footnotesize
\setlength{\tabcolsep}{4pt}
\begin{tabularx}{\columnwidth}{@{}lX@{}}
\toprule
Metric & Value \\
\midrule
Actor & Qwen3-4B-Base \citep{qwen3} \\
Task & \rar{} \citep{rar} \\
Train prompts & 1,600 \\
Seed & 42 \\
RL algorithm & GRPO \citep{grpo} \\
Framework & VERL \citep{verl} \\
Reward variable & Judge model only \\
Responses / prompt & 4 \\
Sampling & Temperature 1, top-$p$ 1 \\
Max prompt length & 1024 \\
Max response length & 1536 \\
Train batch size & 8 \\
PPO mini-batch size & 8 \\
PPO micro-batch size & 1 per GPU \\
PPO epochs / update & 1 \\
Optimizer & AdamW \citep{adamW} \\
Learning rate & $1\times10^{-6}$ \\
Weight decay & 0.01 \\
Gradient clipping & 1 \\
Actor KL coefficient & 0.02 \\
Reward KL & None \\
Training epochs & 5 \\
Save frequency & 200 steps \\
Comparison checkpoint & Step 800 \\
Validation during RL & 50 prompts every 50 steps \\
Final evaluation & 100 reserved prompts \\
Final judge & \gptfour{} \citep{gpt4o} \\
Reward aggregation & Absolute rubric weights \\
\bottomrule
\end{tabularx}
\caption{Shared RL configuration for the controlled \rar{} reward-model comparison.}
\label{tab:rl-config}
\end{table}

\begin{table}[t]
\centering
\small
\setlength{\tabcolsep}{5pt}
\begin{tabular}{lcc}
\toprule
Reward judge & Score & 95\% CI \\
\midrule
None & 0.232 & [0.170, 0.301] \\
Qwen3-0.6B Probe & 0.562 & [0.490, 0.631] \\
Qwen3-1.7B Probe & 0.643 & [0.573, 0.708] \\
Qwen3-4B Probe & 0.588 & [0.514, 0.659] \\
Qwen3-8B Probe & 0.506 & [0.432, 0.575] \\
Qwen3-8B Generative & 0.594 & [0.523, 0.662] \\
\bottomrule
\end{tabular}
\caption{Final \rar{} RL scores with 95\% bootstrap confidence intervals.}
\label{tab:rl-score-ci}
\end{table}

\begin{table*}[!t]
\centering
\small
\setlength{\tabcolsep}{5pt}
\begin{tabularx}{\textwidth}{@{}p{0.18\textwidth}X p{0.26\textwidth} p{0.18\textwidth}@{}}
\toprule
Panel & Measurement & Result & Notes \\
\midrule
Preference counts & Human preference on 100 blind A/B pairs &
Base 6; Probe 72; Tie 13; Unsure 9 &
Probe policy preferred on 72\% of examples. \\
Preference counts & \gptfour{} rubric preference on the same 100 pairs &
Base 9; Probe 80; Tie 11 &
Same candidate pairs and rubric scoring protocol. \\
\midrule
Human-tie analysis & Human-tie subset &
13 / 13 human ties &
Tie labels are analyzed separately from preferences. \\
Human-tie analysis & Small \gptfour{} score gaps among human ties &
7 / 13 have $|\Delta| \leq 0.05$; 11 / 13 have $|\Delta| < 0.20$ &
Most human ties are also near-ties under the rubric judge. \\
\midrule
Preference agreement & Human and \gptfour{} preferences after excluding
human Tie/Unsure labels and \gptfour{} exact ties &
69 / 72 agreement &
Counts: 3 base/base, 2 base/Probe, 1 Probe/base, 66 Probe/Probe. \\
\midrule
Reference-judge check & GPT-5 rescoring on a matched static reference bank &
99 / 100 parsed samples &
Independent reference rescoring uses the same criterion-level target. \\
Reference-judge check & Criterion-level GPT-4o/GPT-5 agreement &
Macro-F1 0.853; item agreement 0.853 &
Agreement is measured over parsed criterion labels. \\
Reference-judge check & Scalar-score GPT-4o/GPT-5 agreement &
Pearson 0.843; Spearman 0.841; MAE 0.125; mean bias -0.004 &
Scores use the same rubric aggregation. \\
\bottomrule
\end{tabularx}
\caption{Human audit and reference-judge validation for the main RL comparison.}
\label{tab:validity-final}
\end{table*}

Table~\ref{tab:rl-final} contains the main controlled RL comparison. The tables below provide implementation details, efficiency measurement details, and supporting validation evidence for that comparison.

\subsection{RL Reward Configuration}
\label{app:rl-implementation-details}

The controlled RL comparisons keep the actor, data, rollout, optimizer, reward aggregation, checkpoint, and final evaluator fixed; the reward judge is the only experimental variable. Table~\ref{tab:rl-config} gives the shared configuration used for the matched rows in Table~\ref{tab:rl-final}.

The generative-judge RL baseline uses the explicit per-criterion protocol. Each criterion is scored by a Qwen3-8B judge generation, parsed into a Yes/No verdict, and aggregated with absolute rubric weights. This is the generated reward used in the main comparison.

The probe reward uses a frozen judge model and a calibrated linear classifier.
Each criterion is rendered in the same canonical format used for static evaluation, scored as a criterion-satisfaction probability, and aggregated with the same absolute-weight rubric rule. The judge model and probe are fixed during RL; only the policy actor is updated.

\subsection{RL Score Confidence Intervals}
\label{app:rl-score-ci}

We compute 95\% bootstrap confidence intervals for final RL evaluation by resampling the 100 reserved \rar{} evaluation prompts with replacement. Table~\ref{tab:rl-score-ci} reports the resulting intervals for the controlled reward-model comparison.

\subsection{On-Policy Reward Dynamics and Qualitative Cases}
\label{app:reward-dynamics}

We examine eight saved step-800 rollout groups for each reward judge, with four candidates per group. We call a group fully saturated when every candidate receives a reward above 0.95. Table~\ref{tab:reward-saturation} shows that the 8B Probe is nearly constant within every sampled group, whereas the selected 1.7B Probe and the 8B Generative judge retain more within-group variation.

\begin{table}[t]
\centering
\scriptsize
\setlength{\tabcolsep}{3pt}
\begin{tabularx}{\columnwidth}{@{}Xrrr@{}}
\toprule
Reward judge & Mean & Group SD & All $>0.95$ \\
\midrule
Qwen3-1.7B Probe & 0.911 & 0.032 & 12.5\% \\
Qwen3-8B Probe & 1.000 & 0.0005 & 100\% \\
Qwen3-8B Generative & 0.706 & 0.102 & 12.5\% \\
\bottomrule
\end{tabularx}
\caption{Step-800 on-policy reward statistics. Group SD is the mean within-group reward standard deviation; the final column is the share of rollout groups in which all four rewards exceed 0.95.}
\label{tab:reward-saturation}
\end{table}

The saturated 8B-Probe rewards provide little ranking signal within a GRPO group, despite the model's strong fixed-bank accuracy. The 1.7B Probe preserves more variation and produces the strongest externally evaluated policy in the controlled comparison. This diagnostic motivates selecting the reward judge by downstream policy quality rather than fixed-bank accuracy alone.

Table~\ref{tab:qualitative-policy-cases} summarizes representative responses from the reserved \rar{} evaluation set. Scores are assigned by the external \gptfour{} evaluator, not by the reward model used for training.

\begin{table}[t]
\centering
\scriptsize
\setlength{\tabcolsep}{3pt}
\begin{tabularx}{\columnwidth}{@{}p{0.21\columnwidth}p{0.25\columnwidth}X@{}}
\toprule
Case & External score & Observation \\
\midrule
Gas partial pressures & Base 0.000; 1.7B Probe 1.000 & The base response repeats definitions; the Probe-reward policy solves both parts. \\
Chemical potential & 1.7B Probe 0.833; 8B Generative $-0.042$ & The Probe-reward policy covers the ensemble relation and sign convention; the Generative-reward policy misses most criteria. \\
Azeotropism & 1.7B Probe 0.167; 8B Generative 0.792 & The Probe-reward policy gives an incorrect mathematical condition, illustrating a substantive failure. \\
8B Probe artifact & 8B Probe 0.000 & The response contains repeated assistant-prefix artifacts despite the saturated internal reward pattern. \\
\bottomrule
\end{tabularx}
\caption{Representative policy outputs from the controlled and diagnostic RL runs. Judge names identify the reward used to train the policy.}
\label{tab:qualitative-policy-cases}
\end{table}

These cases illustrate both improvement and failure modes; they do not establish general robustness to adversarial formatting or reward-seeking behavior.

\subsection{Efficiency Measurement}
\label{app:efficiency-measurement}

The efficiency comparison in Section~\ref{sec:efficiency} is computed from the training traces and validation reward metadata of the matched Qwen3-1.7B probe and Qwen3-8B generative-judge RL runs. The generative judge therefore has 4.7$\times$ as many backbone parameters as the probe judge. For each run, the training traces record per-step reward-judge time and a cumulative judge-time counter.
The cumulative counter includes reward scoring during training and the scheduled validation passes up to the comparison checkpoint. The corresponding per-step training reward totals alone are 8,045.9 seconds for the probe reward and 84,662.4 seconds for the Generative reward, giving a 10.5$\times$ ratio. Including scheduled validation reward scoring yields 8,389.9 seconds and 89,912.1 seconds, respectively, giving the 10.7$\times$ ratio reported in Table~\ref{tab:efficiency-final}.

The validation latency calculation uses the reward metadata from the comparison checkpoint. Both reward models score the same 50 responses, covering 385 total rubric criteria. The probe reward records 31.1 seconds of total judge time (0.6216 seconds per response and 0.0807 seconds per criterion). The generated reward records 492.0 seconds of total judge time (9.8410 seconds per response and 1.2780 seconds per criterion). Both runs have sample-level parse success 1.0 on this validation pass. This parse success does not remove the structural difference between the reward computations: the Generative reward obtains criterion labels by decoding and parsing Yes/No verdicts, while the probe reward obtains criterion probabilities from a frozen hidden-state classifier.

The elapsed training times in Table~\ref{tab:efficiency-final} come from the progress traces at the comparison checkpoint. Their ratio is smaller than the cumulative judge-time ratio for two reasons. First, the judge-time counter is an aggregate over reward calls, and those calls can run concurrently. Second, actor rollout, reference-model scoring, optimization, synchronization, and checkpointing remain in the critical path for both runs. The elapsed time still decreases from 11h 11m 56s to 8h 51m 47s through the comparison checkpoint, a 2h 20m 09s reduction. We therefore use cumulative judge time as the primary efficiency metric for the reward model and report elapsed training time as the full-pipeline runtime effect.

\subsection{Human Audit and Reference-Judge Validation}
\label{app:human-audit}

The human audit evaluates whether the main RL improvement is visible beyond the \gptfour{} rubric score. We compare Qwen3-4B-Base against the Qwen3-1.7B probe-reward policy on 100 reserved \rar{} examples. The audit is model-name blind with randomized A/B order. For each example, the annotator sees the question, reference answer, rubric titles, and the two candidate responses, then marks pairwise preference and optional artifact notes.

The audit favors the probe-reward policy on 72 examples, the base actor on 6, with 13 ties and 9 unsure labels. Among the 13 human ties, 7 have an absolute \gptfour{} score gap at most 0.05 and 11 have a gap below 0.20 (Table~\ref{tab:validity-final}), indicating that many human ties are also near-ties under the rubric judge. When human Tie/Unsure labels and \gptfour{} exact ties are excluded, the human preference and \gptfour{} rubric preference agree on 69 of 72 examples (Table~\ref{tab:validity-final}). These results are used as supporting validation evidence for the operational \gptfour{} target, not as supervised labels for training the reward model.

Table~\ref{tab:validity-final} also reports a reference-judge check using GPT-5 \citep{gpt5} on a matched static reference bank.

\subsection{Additional RL Configuration Checks}
\label{app:rl-configuration-checks}

During development, we ran additional RL checks that varied KL coefficient, maximum response length, checkpoint choice, and probe calibration size. These runs were used to choose a stable matched comparison protocol. We do not report them as competing results because they vary training conditions rather than isolating the reward judge. The main paper therefore reports only the controlled comparison in Table~\ref{tab:rl-final}, where actor, data, GRPO configuration, checkpoint, response length, reward aggregation, and \gptfour{} evaluation are fixed across reward models.

\section{Generalization Evaluation Details}
\label{app:generalization-protocols}

\subsection{\gpqa{} Evaluation}
\label{app:gpqa-protocol}

The \gpqa{} evaluation uses the GPQA-Diamond split with 198 questions \citep{gpqa}. Each example is converted into a four-choice prompt. The answer choices are permuted across four runs, and the model is asked to place the final answer letter inside \texttt{\textbackslash boxed\{\}}. Accuracy is computed against the permuted gold answer. When the boxed-answer parser fails, \gptfour{} \citep{gpt4o} is used only as a fallback verifier for the final choice. Table~\ref{tab:gpqa-transfer-final} gives the mean and standard deviation across the four answer-order runs.

\subsection{Cross-Domain Probe Transfer}
\label{app:cross-domain-protocol}

For cross-domain transfer, a Qwen3-1.7B linear probe \citep{qwen3} is fit on source-domain criterion labels and evaluated on target-domain examples without updating the probe on target-domain labels. Each source domain uses 350 training questions and 50 development questions for layer and threshold selection. Evaluation uses 100 target-domain questions scored by \gptfour{} \citep{gpt4o} under the same pointwise criterion-satisfaction target used in the static experiments. The metrics in Table~\ref{tab:cross-domain-final} compare the probe predictions with \gptfour{} criterion labels and aggregated rubric scores.

\raggedbottom

\section{Prompts for Reproducibility}
\label{app:prompts}

This appendix lists the prompt templates used for response construction and criterion-level rubric scoring. Bracketed fields are replaced with the corresponding dataset values. For local Qwen judges, the user prompt is wrapped with the model chat template before inference. The reference rubric-scoring prompt is used for \gptfour{} labels; generative, logprob, and probe judges use the itemwise criterion prompts.

\begin{promptboxcol}{Generative Criterion Scoring}
You are evaluating whether an AI response satisfies a specific rubric criterion.

[Question]
{question}

[Criterion to Evaluate]
Title: {title}
Description: {description}

[AI Response]
{response}

Evaluate whether the AI response satisfies this criterion.
Output exactly one final line and no explanation:
- 1 if the criterion is satisfied
- 0 if the criterion is NOT satisfied

Your entire response must be exactly one of:
\boxed{{0}}
\boxed{{1}}
\end{promptboxcol}

\begin{promptboxcol}{Reference Rubric Scoring}
You are an impartial rubric grader.

Instruction:
{question}

Candidate Response:
{response}

Rubric Rules ({num_rules} total):
{rubric_rules}

Task:
For each rubric rule, decide whether the response satisfies it.

Output format requirement:
- Return exactly {num_rules} lines.
- Each line must follow:
  Rule i: <short justification>. Verdict: Yes/No
- Use i = 1..{num_rules} in order.
- Verdict must be exactly `Yes` or `No`.

Do not output JSON, markdown, extra headings, or any text outside those rule lines.
\end{promptboxcol}

\begin{promptboxcol}{Forced-choice Logprob Scoring}
You are an impartial rubric grader.

Instruction:
{question}

Candidate Response:
{response}

Rubric Criterion:
{criterion}

Question:
Does the candidate response satisfy this criterion?

Answer with exactly one word: Yes or No.

Answer:
\end{promptboxcol}

\begin{promptboxcol}{Probe Representation Scoring}
You are an impartial rubric grader.

Instruction:
{question}

Candidate Response:
{response}

Rubric Criterion:
{criterion}

Question:
Does the candidate response satisfy this criterion?

Answer:
\end{promptboxcol}

\begin{promptboxcol}{\rar{} Candidate Response Generation}
{question}
\end{promptboxcol}

\begin{promptboxcol}{\orb{} Full-Score Response Generation}
You are a highly knowledgeable and versatile AI assistant. Your goal is to provide accurate, well-structured, and helpful answers.

For the following question, follow these requirements:
1. First, think step-by-step and provide your internal reasoning.
2. Then, provide a clear and concise final answer.
3. Your answer must satisfy each criterion in the rubric.

Question: {instruction}

Rubric: {rubric}

--- YOUR ANSWER ---
### Reasoning:
[Your step-by-step reasoning here]

### Final Answer:
[Your final answer here]
\end{promptboxcol}

\begin{promptboxcol}{\orb{} Partial-Compliance Response Generation}
You are a selective and inconsistent AI assistant. Your goal is to provide a response that follows around half of the rubric rules while deliberately ignoring others.

For the following question, first decide which criteria to obey and which to disobey. Then provide a final answer that clearly satisfies the chosen criteria while noticeably failing the others.

Question: {instruction}

Rubric: {rubric}

--- YOUR ANSWER ---
### Rules you decide to obey:
[List what rules you obeyed and what you disobeyed]

### Reasoning:
[Your reasoning here]

### Final Answer:
[Your final answer here]
\end{promptboxcol}

\begin{promptboxcol}{\orb{} Low-Score Response Generation}
You are a deceptive and subtle AI assistant. Your goal is to provide an answer that appears authoritative but is irrelevant to the actual question and fails the rubric.

Question: {instruction}

Rubric: {rubric}

--- YOUR ANSWER ---
### Reasoning:
[Your flawed reasoning here]

### Final Answer:
[Your final answer here]
\end{promptboxcol}

\begin{promptboxcol}{\orb{} Unguided Response Generation}
Question: {instruction}
Your Answer:
\end{promptboxcol}

\begin{promptboxcol}{\orb{} Rubric Standardization}
You are a precise and analytical editor. Refine and reorganize the given rubric into a standardized JSON format.

Input:
Question ID: {question_id}
Original Rubric:
{rubric}

Rules:
1. Rules marked `[Hard Rule]` must be type `"hard"`.
2. Rules marked `[Principle]` must be type `"soft"`.
3. Output exactly 6 rules: 2 hard and 4 soft.
4. Merge or split rules as needed while preserving the original intent.
5. Each refined rule should be clear, objective, and concise.

Output only a JSON array:
[
  {
    "question_id": "{question_id}",
    "rule_id": "1",
    "type": "hard" or "soft",
    "rubric": "..."
  }
]
\end{promptboxcol}

\flushbottom
\section{Dataset Examples}
\label{app:dataset-example}

The examples below show the pointwise labels used in the two data settings. We choose responses that are useful but imperfect, so the examples show how the rubric separates high-quality partial compliance from full satisfaction.

\paragraph{\orb{} example.}

\textbf{Prompt.} How do I check if a new install of a version is different from the current one, booting from SBOM?

\textbf{Candidate response.} To determine if a new installation version is different from the current one by comparing SBOMs, obtain the SBOM for the current installed version, generate the SBOM for the new installation, and compare the component lists. Focus on package names, versions, checksums, added or removed packages, dependency changes, and licensing changes. The response also recommends tools such as Syft, CycloneDX, and ORT, and suggests automating the comparison in a deployment pipeline.

\textbf{\gptfour{} score.} 9 / 10.

\textbf{Verdicts.} Yes, Yes, Yes, No, Yes, Yes.

\begin{table}[t]
\centering
\scriptsize
\setlength{\tabcolsep}{3pt}
\begin{tabularx}{\columnwidth}{@{}p{0.18\columnwidth}Xc@{}}
\toprule
Item & Criterion summary & Verdict \\
\midrule
Hard 1 & Outlines a methodical comparison process using an SBOM. & Yes \\
Hard 2 & Uses the SBOM as part of the comparison procedure. & Yes \\
Soft 3 & Provides multiple approaches with concrete examples or commands. & Yes \\
Soft 4 & Uses clear labeled organization and identifies limitations. & No \\
Soft 5 & Recommends appropriate specialized tools or techniques. & Yes \\
Soft 6 & Explains how to interpret differences and verify assumptions. & Yes \\
\bottomrule
\end{tabularx}
\caption{Criterion-level labels for the \orb{} example.}
\label{tab:orb-example-verdicts}
\end{table}

\paragraph{\rar{} example.}

\textbf{Prompt.} An ideal gas at NTP has a most probable velocity $\bar{u}=4.20\times 10^4\,\text{cm/sec}$ and a mean free path of $7.90\times 10^{-8}\,\text{cm}$. Calculate the mean time between collision.

\textbf{Candidate response.} The response uses $\tau=\lambda/\bar{u}$ and substitutes the given mean free path and most probable velocity: $\tau=(7.90\times 10^{-8})/(4.20\times 10^4)\approx 1.88\times 10^{-12}$ seconds. The response is clear and numerically consistent, but it does not convert the most probable velocity to the average velocity before applying the mean-collision-time formula.

\textbf{\gptfour{} score.} 0.773.

\textbf{Verdicts.} Yes, No, Yes, Yes, Yes, Yes, Yes.

\begin{table}[t]
\centering
\scriptsize
\setlength{\tabcolsep}{3pt}
\begin{tabularx}{\columnwidth}{@{}p{0.25\columnwidth}Xc@{}}
\toprule
Item & Criterion summary & Verdict \\
\midrule
Formula & Applies $\tau=\lambda/\nu$ for mean time between collisions. & Yes \\
Velocity & Converts most probable velocity to average velocity using $2/\sqrt{\pi}$. & No \\
Substitution & Substitutes the given numerical values accurately. & Yes \\
Units & Keeps velocity and length units consistent and reports seconds. & Yes \\
Derivation & Presents a clear derivation from relation to final value. & Yes \\
Precision & Uses appropriate significant figures in the final answer. & Yes \\
Assumptions & Does not introduce extraneous non-ideal effects. & Yes \\
\bottomrule
\end{tabularx}
\caption{Criterion-level labels for the \rar{} example.}
\label{tab:rar-example-verdicts}
\end{table}